\documentclass[preprint,11pt,authoryear]{elsarticle}

\usepackage{amssymb}
\usepackage{amsmath}
\usepackage{graphicx}
\usepackage{pdfpages}
\usepackage{graphicx,subfigure}
\usepackage[plain,noend]{algorithm2e}
\usepackage{algorithmic}
\usepackage{comment}
\usepackage{caption}
\usepackage{textcomp}
\usepackage[graphicx]{realboxes}
\usepackage{multirow}
\usepackage{tikz}
\usepackage{textcomp}
\journal{European Journal of Operational Research}

\begin{document}

\includepdf[pages=-]{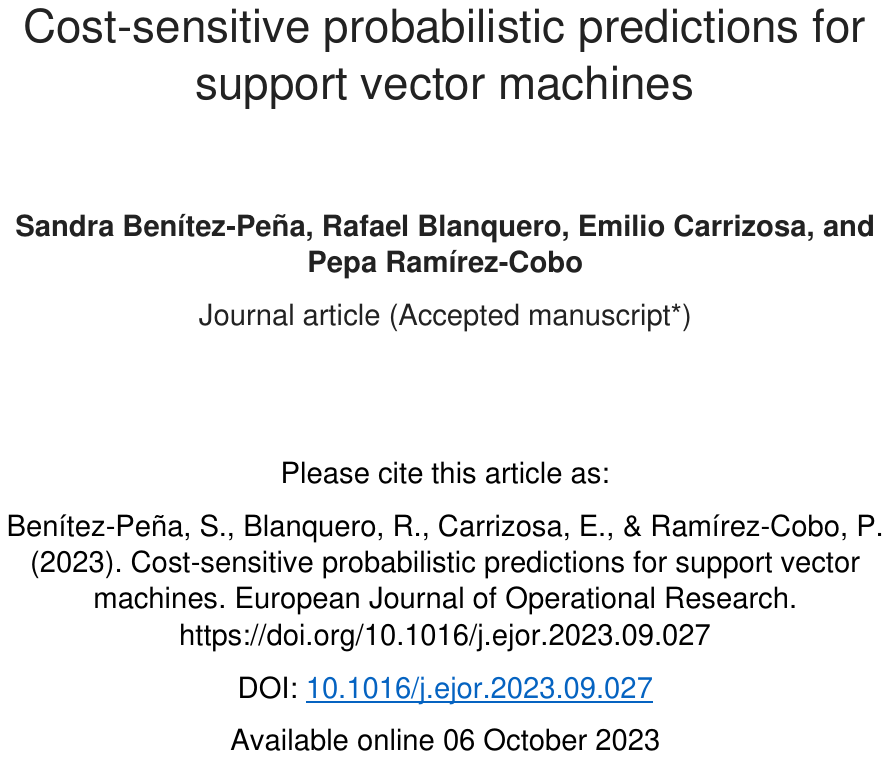}

\begin{frontmatter}



\title{Cost-sensitive probabilistic predictions for support vector machines}


\author[a]{Sandra Benítez-Peña}
\author[b,c]{Rafael Blanquero}
\author[b,c]{Emilio Carrizosa}
\author[b,d]{Pepa Ramírez-Cobo}

\affiliation[a]{organization={Department of Statistics, University Carlos III of Madrid},%Department and Organization
            %addressline={},
            %city={},
            %postcode={},
            %state={},
            country={Spain}}

\affiliation[b]{organization={Instituto de Matem\unexpanded{á}ticas de la Universidad de Sevilla},%Department and Organization
            %addressline={},
            %city={},
            %postcode={},
            %state={},
            country={Spain}}

\affiliation[c]{organization={Department of Statistics and Operations Research, University of Sevilla},%Department and Organization
            %addressline={},
            %city={},
            %postcode={},
            %state={},
            country={Spain}}

\affiliation[d]{organization={Department of Statistics and Operations Research, University of C\unexpanded{á}diz},%Department and Organization
            %addressline={},
            %city={},
            %postcode={},
            %state={},
            country={Spain}}

\begin{abstract}

Support vector machines (SVMs) are widely used and constitute one of the best examined and used machine
learning models for two-class classification. Classification in SVM is based on a score procedure, yielding a deterministic classification rule, which can be transformed into a probabilistic rule (as implemented in off-the-shelf SVM libraries),
but is not probabilistic in nature. On the other hand, the tuning of the regularization parameters
in SVM is known to imply a high computational effort and generates pieces of information that are not fully exploited, not being used to build a probabilistic classification rule.

\textcolor{black}{In this paper we propose a novel approach to generate probabilistic outputs for the SVM. The new method has the following three properties. First,
it is designed to be cost-sensitive, and thus the different importance of sensitivity (or true positive rate, TPR) and specificity (true negative rate, TNR) is readily accommodated in the model. As a result, the model can deal with imbalanced datasets which are common in operational business problems as churn prediction or credit scoring.  Second, the SVM is embedded in an ensemble method to improve its performance, making use of the valuable information generated in the parameters tuning process. Finally, the probabilities estimation is done via bootstrap estimates, avoiding the use of parametric models as competing approaches.
%\textcolor{red}{\sout{This allows us to compute confidence intervals for the score and class probabilities for a given individual.}}
Numerical tests on a wide range of datasets show the advantages of our approach over benchmark procedures.}

\end{abstract}



\begin{keyword}

Machine Learning \sep Support Vector Machines \sep Probabilistic Classification \sep Cost-Sensitive Classification.



\end{keyword}

\end{frontmatter}



\section{Introduction}\label{Intro}

Supervised classification is one of the most relevant tasks in Data Science. We are given
a set $\Omega$ of individuals. Each element $i \in \Omega$ is represented by a pair
$(x_i,y_i)$, where ${x}_i \in \mathbb{R}^n$ is the attribute vector, and $y_i \in
\mathcal{C}$ is the class membership of object $i$. We only have class information in $T \subset
\Omega$, which is called the \textit{training sample}. In its most basic version, the one considered in this paper, supervised
classification addresses two-class problems, that is to say,  $\mathcal{C} = \{-1,+1\}$.

Support Vector Machine (SVM) is a powerful and state-of-the-art method in supervised
classification that aims, in the simplest case of linear SVM, at separating both classes by means of a linear classifier, $\omega^{\top}
x_i + \beta$. The coefficients $\omega, ~ \beta$ of the SVM can be obtained by solving a
convex quadratic programming (QP) formulation with linear constraints. It is usual to consider its dual formulation, which allows us to use the so-called kernel trick (\cite{vapnik1998statistical, hofmann2006support}), and is given by

\begin{equation}
\label{eq:svm}
\begin{array}{lll}
\max\limits_{\mathbf{\alpha}} & -\dfrac{1}{2}\sum_{i \in T}\sum_{j \in T} \alpha_i \alpha_j y_i y_j K(x_i,x_j) + \sum_{i \in T} \alpha_i &\\
s.t. & \sum_{i \in T} \alpha_iy_i = 0 & \\
& 0 \leq \alpha_i \leq C, & i \in T\\
\end{array}
\end{equation}
%\begin{equation}
%\label{eq:svm}
%\begin{array}{lll}
 % \min_{\mathbf{\omega}, \beta, \mathbf{\xi}} & %\mathbf{\omega}^\top \mathbf{\omega} + C\sum_{i \in
  %I} \xi_i& \\
  %s.t. & y_i(\mathbf{\omega}^\top {x}_i + \beta) \geq 1 - \xi_i,& i \in T \\
   %& \xi_i \geq 0& i \in T,\\
  % & (\omega,\beta) \in \Omega &
%\end{array}
%\end{equation}
where $\alpha$ are the usual variables of the dual SVM formulation, $C>0$ is a \textit{regularization parameter} to be tuned, which controls the
trade-off between margin maximization and misclassification errors, and $K$ is a \textit{kernel} such that $K(x_i,x_j) = \phi(x_i)^\top\phi(x_j)$ (where $\phi$ is a mapping function that embeds the dataset into a higher dimensional space). The kernel function $K$ may include other parameters, such as the $\sigma$ parameter in the Radial Basis Function (RBF) kernel (see e.g., \cite{herbrich2001learning, hofmann2008kernel}). Such parameters have to be also tuned, in a grid in values {which we will denote, for simplicity's sake, as} $\theta \in \Theta$. For further details, see e.g. \cite{carrizosa2013supervised} and references therein.

Given an object \textcolor{black}{with attribute vector $x_0$,} the SVM algorithm produces a hard labeling in such a way that this instance is classified in the positive or
the negative class according to the sign of $f({x}_0)$, where
$f(x)=\sum_{i \in T} \alpha_i y_i K(x,x_i) + \beta$ is the \textit{score function}. {When an attribute vector $x_0$ is given, the value $f(x_0)$ is called the \textit{score value} of $x_0$.}  However, the SVM method does not
result in probabilistic outputs as posterior probabilities $P(y=+1\mid x)$, which are of interest if
a measure of confidence in the predictions is sought, see \cite{Murphy}.
%This is of particular importance in several real-world applications such as cancer screening or credit scoring, where the risk of a false-negative and a false-positive results are significantly different.
\textcolor{black}{This is of particular importance in several business problems such as churn prediction, e.g., \cite{huang2012customer}, to calibrate the probability of churning of a customer, or credit scoring, e.g., \cite{thomas2017credit}, where the probability of defaulting is to be estimated. In these situations, it is important not only to get a hard label for the individual, but also an estimation of the degree of confidence in the assignation.}

\textcolor{black}{Several attempts {to obtain the posterior probabilities $P(y=+1\mid x)$ for SVM} have been already carried out previously. One of
them is based on assigning posterior class probabilities assuming a specific parametric family for the posterior probability}. For
example, \cite{wahba1992multivariate}, \cite{wahba1999support} proposed a logistic link function,

\begin{equation}\label{Wahba}
  P(y=+1\mid x) = \dfrac{1}{1+exp(-f(x))}.
\end{equation}

Also, \cite{vapnik1998statistical} suggested to estimate $P(y=+1\mid x)$ in terms of a series of the
trigonometric functions, where the coefficients of the trigonometric expansion minimizes a regularized function. \textcolor{black}{Another considered option has been to fit Gaussians} to the class-conditional densities $P(f(x) \mid y=+1)$ and
$P(f(x) \mid y=-1)$, as proposed in \cite{hastie1998classification}. \textcolor{black}{From such a choice}, the
posterior probability $P(y=+1 \mid x )$ is assumed to be a sigmoid, whose slope is determined by the tied variance.
%Hastie and Tibshirani adjusted this probability so that $P(y=1|f(x)) = 0.5$ happened at $f(x) = 0$.
One of the best-known heuristics to obtain probabilities is due to
\cite{platt1999probabilistic}, which considers $f(x)$ as the log-odds ratio $\log \dfrac{P(y=+1\mid
x)}{P(y=-1\mid x)}$. This implies that
\begin{equation}\label{Platt}
  P(y=+1 \mid x) = \dfrac{1}{1+\exp(A f(x)+B)},
\end{equation}
and $A$ and $B$ can be estimated by maximum likelihood on a validation set. This technique is
implemented by well-known statistical packages such as the {\tt ksvm()} function in R, see \cite{karatzoglou2006}\textcolor{black}{, } {\tt predict\_proba} in scikit-learn in Python (\cite{pedregosa2011scikit}) \textcolor{black}{ or in the software LIBSVM (see \cite{LIBSVM}), which uses a better implementation of the method, as presented in \cite{lin2007note}. Although SVM is designed for binary classification, there are several extensions for multiclass problems, e.g. \cite{carrizosa2008multi, lorena2008evolutionary, wang20071}, and also some attempts to construct class probabilities are found in the literature. In particular, multiclass versions \textcolor{black}{of Platt's approach} can be found in \cite{1556171} and \textcolor{black}{have been implemented in software packages} like LIBSVM (\cite{LIBSVM})}.

\begin{figure}
  \centering
  \includegraphics[width=10cm]{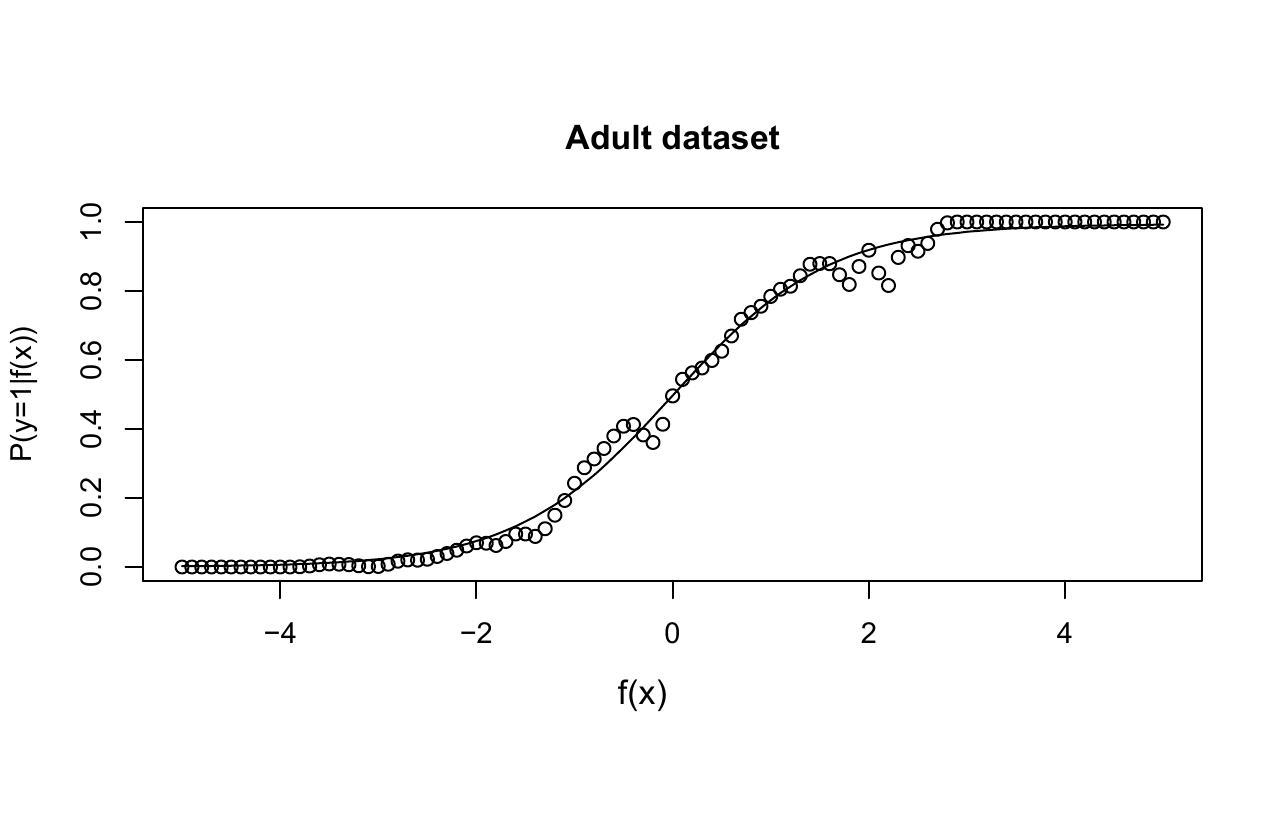}
  \includegraphics[width=10cm]{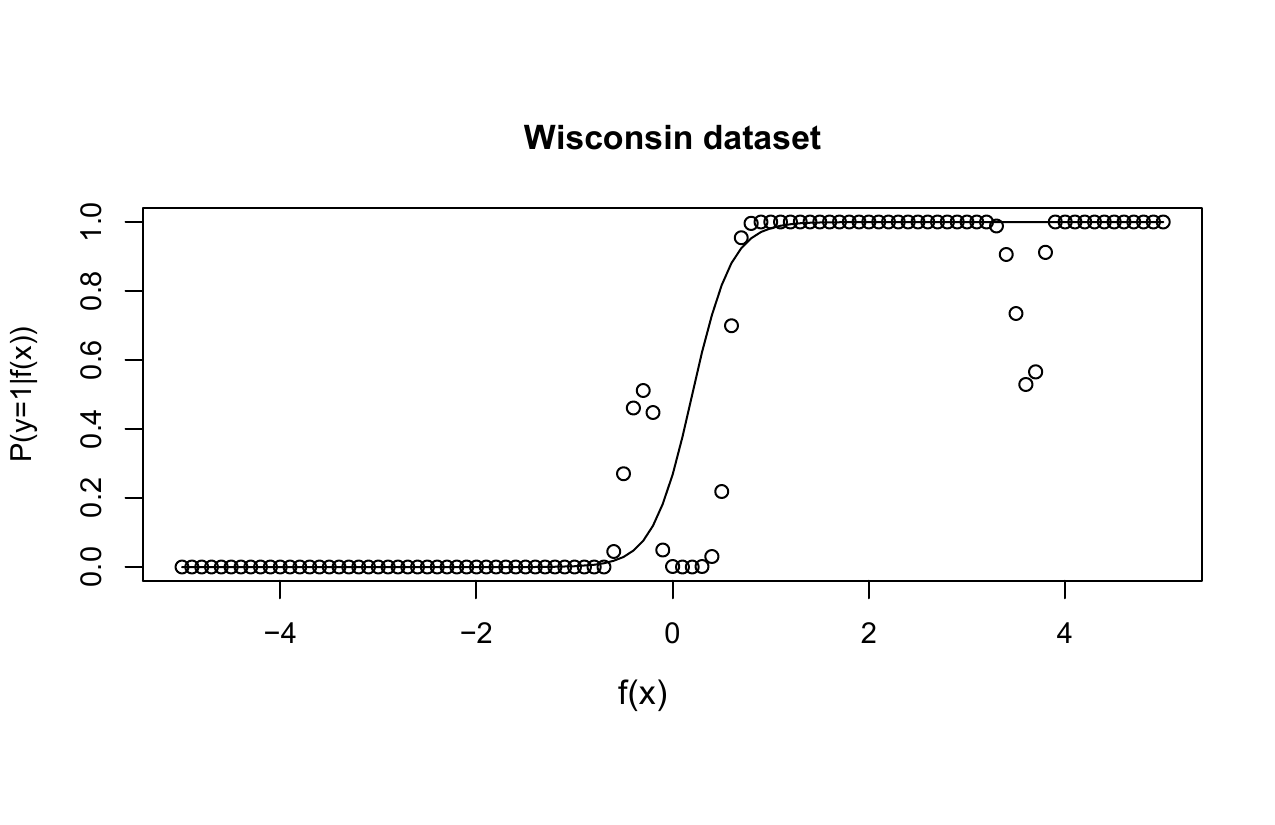}
  \caption{Fit (solid line) of the sigmoid function to the empirical class probabilities (dots) of \texttt{adult} and \texttt{wisconsin} datasets.}\label{fig:PlattProced}
\end{figure}

\textcolor{black}{Platt's approach} has been criticized for failing
to provide insight and for interpreting $f(x)$ as a log-odds ratio, which may be not accurate for some
datasets, see \cite{Murphy,tipping2001sparse,Franc2011}. To illustrate such a phenomenon, consider
Figure \ref{fig:PlattProced}, which shows the fit of the sigmoid function (\ref{Platt}) to the empirical class probabilities of two different, well-referenced datasets: \texttt{adult} and \texttt{wisconsin}, respectively (see Section~\ref{subsec:DataExp}). It can be seen that, while for \texttt{adult} dataset the fit provided by the method given in (\ref{Platt}) performs reasonably well, the performance is poor for \texttt{wisconsin}.

 \cite{sollich2002bayesian} considers a different probabilistic framework for SVM classification, based on Bayesian theory. In particular, it relates the SVM kernel to the covariance function for a Gaussian process prior and, as a result, optimal values of the tuning parameter $C$
and class probabilities are obtained in a natural way. \textcolor{black}{Again, this method, as the previously commented approaches, make modeling assumptions that might not be satisfied by the data}. Finally, other procedures seeking probabilistic outputs are found in the literature, as
\cite{seeger2000bayesian}, \citet{kwok1999moderating,kwok1999integrating}, \cite{herbrich1999bayesian}. %

%Even though

\textcolor{black}{None of the previously mentioned works produce cost-sensitive models, which are of crucial importance in many managerial decision-making problems. For instance, in a churn prediction context, classifying a churned customer as non-churned may have important negative consequences. In a similar way and in order to avoid high costs, it is more important for a financial institution to correctly classify a defaulting customer than a non-defaulting one. Comparable situations arise in other settings different from business and management domains as medical diagnosis, in which failing to detect a disease may have fatal aftermaths. Because of that, cost-sensitive classification has become a trending issue lately and a number of references can be found regarding this, see \cite{ARAM2022116683, VANDERSCHUEREN2022400,MALDONADO2021102380,DEBOCK2020612,coussement2014improving,bradford1998pruning,freitas2007cost,carrizosa2008multi,datta2015near,Sandra1,Sandra2}.}

\textcolor{black}{Cost-sensitivity is closely related to the problem of imbalancedness in datasets. Imbalancedness may produce innacurate classification rates for the minority class that is often the most critical one, \cite{app10072581}. Several attempts in the literature have considered probabilistic outputs for the SVM in a context of imbalancedness. For example, \cite{tao2005posterior} propose robust SVM that turn out insensitive to the class imbalancedness.} Their approach, the \textit{posterior
probability support vector machine} (PPSVM), is distribution-free and weighs imbalanced training samples. A multiclass approach based on the method in \cite{tao2005posterior} is proposed by \cite{4359207}. Also, a more sophisticated and computationally expensive alternative is proposed by \cite{KIM201519}, which combines layers of SVM with class probability output networks (CPONs), in which strong statistical assumptions are imposed.

\textcolor{black}{In this work we provide a {non-parametric method (and therefore, contrary to some of the competitors, free of any assumptions about our data)} for obtaining point estimates for the class probabilities for the SVM which addresses properly cost sensitivity, since the rates of main interest (either TPR or TNR, which are the probabilities of an individual with label $y=+1$ and $y=-1$ respectively, being classified in class $+1$ and $-1$, respectively) are explicitly controlled. As an example, and continuing with the credit scoring problem, we consider the dataset \texttt{german}, available at the UCI Repository \citep{Dua2017} and described in detail in Section~\ref{sec:ER}. This dataset is slightly imbalanced (the class of defaulting customers represents the $30\%$ of the total). The mean squared error (MSE) of the probability prediction for  the defaulting class under the non cost-sensitive version of the novel method  is equal to $0.51$ (average value over a testing sample). If we use instead the cost-sensitive version, this error decreases down to $0.134$. In the setting of churn prediction we obtain similar results. Again, the considered database \texttt{churn} is imbalanced, the percentage of churners being equal to $15.71\%$ (see Section~\ref{sec:ER}). From an initial mean squared error equal to $0.8$ the new model is able to decrease it down to $0.149$. As it will be commented, such reductions may be at the expense of damaging the prediction of the posterior negative class probabilities, which are assumed to be less relevant.}

\textcolor{black}{Another distinctive feature of our approach is that the SVM is embedded in an ensemble method {(see \cite{petrides2022cost} and references therein for more details regarding ensemble methods)} which, as will be shown, means an improvement in performance {(\cite{DEBOCK20126816,wang2009empirical,BENITEZPENA2021648})}.} It is known that, in order to solve the SVM problem (\ref{eq:svm}), a tuning process concerning the regularization parameters in the grid $\Theta$ needs to be performed. Traditionally, all the information resulting from this tuning procedure is discarded and only the \textit{best} value $\theta \in \Theta$ is used to build the classifier. Instead, in this work, the final posterior class probability estimate is a weighted mean of different posterior probabilities, each one related to specific values of $\Theta$.
In addition, here we propose a novel methodology that does not make use of parametric models based on the \textcolor{black}{score function $f(x)$} obtained after tuning the SVM parameters. Instead, we consider a bootstrap framework (\cite{efron1986bootstrap, efron2000bootstrap}), which, to the best of our knowledge, has not been addressed before for this type of problems. The use of a bootstrap sampling allows us to obtain accurate values for the density of the score values, which translates into a better prediction of the posterior class probability {$P(y=+1\mid x)$}. {As we have already seen from Figure~\ref{fig:PlattProced}, considering certain parametric assumptions can lead to poor fitting results of the probabilities. By considering a non-parametric approach as we do here, we do not presume assumptions that in some cases may be unrealistic. On the other hand, as it will be shown in detail both in Section~\ref{sec:ER} and in the Supplementary Material, our method turns out more flexible and leads to a better fit of the real probabilities for many cases.} %Furthermore, \textcolor{black}{confidence intervals for both the score values and the estimated class probabilities can be obtained this way. This would provide a higher confidence to our results \textcolor{black}{(\cite{DEBOCK20126816})} and it is another novelty of our approach with respect to off-the-shelf procedures in the literature, although further research is necessary here. The construction of confidence intervals will be object of study in a future work.}
%\textcolor{black}{
%As a particular example of all these novel contributions, we refer to the different results we have obtain for \texttt{german} credit scoring dataset. First, we have calculated the overall MSE for the estimated probabilities, obtaining similar results to standard approaches. Second we can control the MSE in the bad customers class, going from an error of $0.510$ when nothing is imposed to a MSE of $0.175$ in this class when a correct classification rate of $0.7$ is imposed, without damaging too much the other class (MSE $0.145$). Furthermore, we have represented the commented confidence intervals for some instances in Figure~\ref{fig:6}.
%}

%la estructura hay que cambiarla según hablamos antes antes de irnos de vacaciones.
The paper is structured as follows. First, in Section~\ref{sec:CSPP}, \textcolor{black}{our} methodology is introduced. Section~\ref{subsec:BootsProp} describes how to integrate a bootstrap sampling into a SVM to enhance accuracy and to produce posterior class probabilities estimates. Section~\ref{subsec:CSSM} explains two different ways to obtain cost-sensitive probabilistic predictions. In Section~\ref{sec:ER} some experimental results are
presented. In particular, several well-referenced datasets from \textcolor{black}{business, social sciences and other} contexts are analyzed. Estimates of the posterior class probabilities under \textcolor{black}{our} methodology are compared to those obtained under benchmark approaches. Finally, the posterior probabilities of the classes of interest are controlled via the two different approaches described in Section~\ref{subsec:CSSM}. Conclusions and further research can be found in Section~\ref{sec:Conc}.

\section{{Cost-sensitive predictive probabilities for SVM}}\label{sec:CSPP}

In this section we present our methodology to obtain {point estimates for the posterior class  probabilities using the SVM classifier {together with a bagging procedure (\cite{wang2009empirical, 101007354045665131})}.} First, in Section~\ref{subsec:BootsProp} we explain how to integrate a bootstrap sampling into the SVM to produce posterior class probabilities estimates \textcolor{black}{$P(y=+1\mid x)$}. Second, in Section~\ref{subsec:CSSM} we describe two different approaches that allow us to control the posterior probability estimates in case we have a binary classification problem with one of the classes of most interest.

\subsection{Bootstrap estimation on posterior class probabilities}\label{subsec:BootsProp}

\begin{figure}
\centering     %%% not \center
\subfigure[]{\label{fig:3a}\includegraphics[width=60mm]{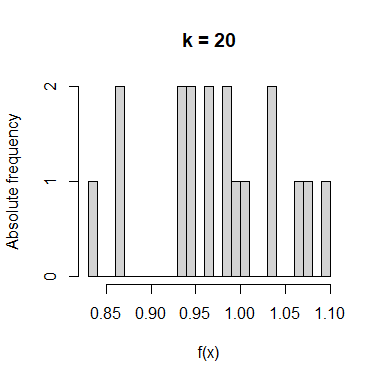}}
\subfigure[]{\label{fig:3b}\includegraphics[width=60mm]{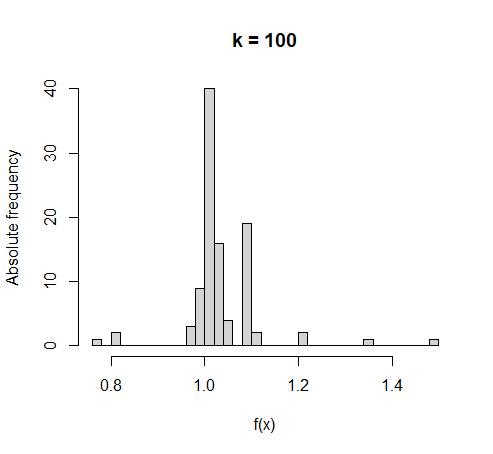}}
\subfigure[]{\label{fig:3c}\includegraphics[width=60mm]{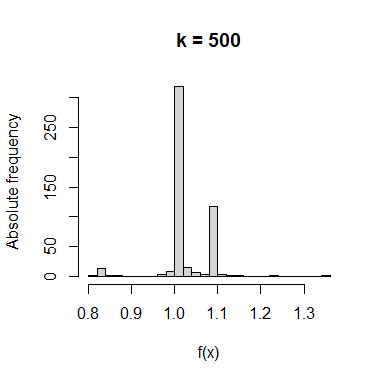}}
\caption{Histogram of scores for a single instance {from the test sample} when a $k$-fold CV is used. Here, we set $k=20$, $100$ and $500$, obtaining as many score values as the value of $k$. {The $x$-axis represents the obtained scores, while the $y$-axis shows the absolute frequency. In each histogram, the sum of the absolute frequencies must be equal to $k$.}}\label{figure2}
\end{figure}

\begin{figure}
\centering     %%% not \center
\subfigure[]{\label{fig:4a}\includegraphics[width=60mm]{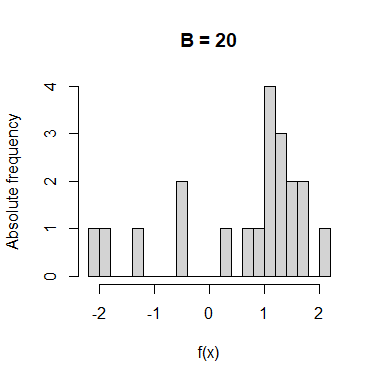}}
\subfigure[]{\label{fig:4b}\includegraphics[width=60mm]{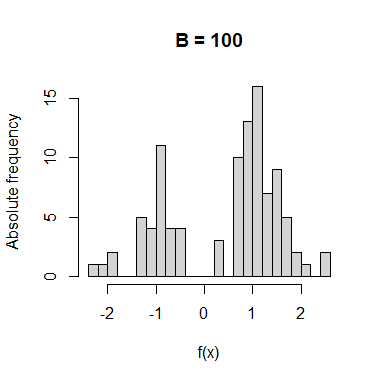}}
\subfigure[]{\label{fig:4c}\includegraphics[width=60mm]{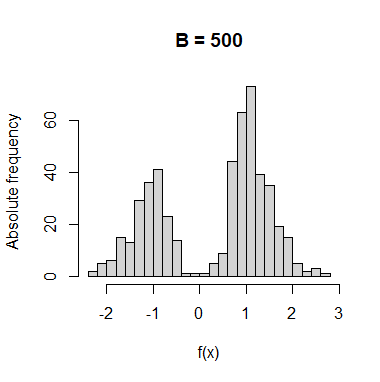}}
\caption{Histogram of scores for a single instance when the Bootstrap with $B$ replications is used. As in the $k$-fold CV, here we set $B=20$, $100$ and $500$. {The $x$-axis represents the obtained scores, while the $y$-axis shows the absolute frequency. In each histogram, the sum of the absolute frequencies must be equal to $B$.}}\label{figure3}
\end{figure}

Assume that we want to solve SVM (\ref{eq:svm}) to classify the observations in a dataset. In order to estimate the classification error of the SVM classifier, it is standard to consider a $k$-fold cross-validation (CV), see \cite{kohavi1995study}. Figure~\ref{figure2} shows the histogram (in absolute frequencies) of the score values under three different choices of $k$ ($k=20, 100, 500$) for a given {instance randomly chosen from the test sample. Thus, the sum of these absolute frequencies must be equal to $k$}.
It can be observed that as $k$ increases, the score values are less disperse, a consequence of the fact that the different samples share more elements, and thus they yield more similar scores (since the support vectors, the elements that {actually} define the hyperplane, are almost the same in each fold). Thus, in this setting the
$k$-fold cross validation strategy does not turn out
convenient since the  estimated probabilities
highly depend on the value of $k$. In particular,
as $k$ increases, the estimated values get
close to $0$ or $1$. What it is proposed in this paper is to replace the $k$-fold CV approach by a bootstrap sampling {which consists in obtaining a random sample with replacement from the original sample}. As it will be shown,
this will allow us to avoid the degenerate behaviour observed in Figure~\ref{figure2}. The {resulting scores} are those illustrated in Figure~\ref{figure3}, where the analogous histograms to Figure~\ref{figure2} are shown, but where a bootstrap sampling with $B$ replications ($B = 20,100,500$) has been considered instead.
\textcolor{black}{The idea of using those values is just to illustrate the behavior of this method when increasing the sample size in contrast with the one shown in Figure~\ref{figure2}}. Finally, as already exposed, the estimates for the posterior class probabilities \textcolor{black}{$P(y=+1\mid x)$} will be obtained as the \textcolor{black}{relative frequency} of the positive (negative \textcolor{black}{in the case of $P(y=-1\mid x)$}) score values. %This approach, given by Algorithm~\ref{algBoot} and illustrated as a flowchart in Figure~\ref{fig:bootstrapflow}, is detailed next.

\begin{figure}
  \centering
  \includegraphics[width=12cm]{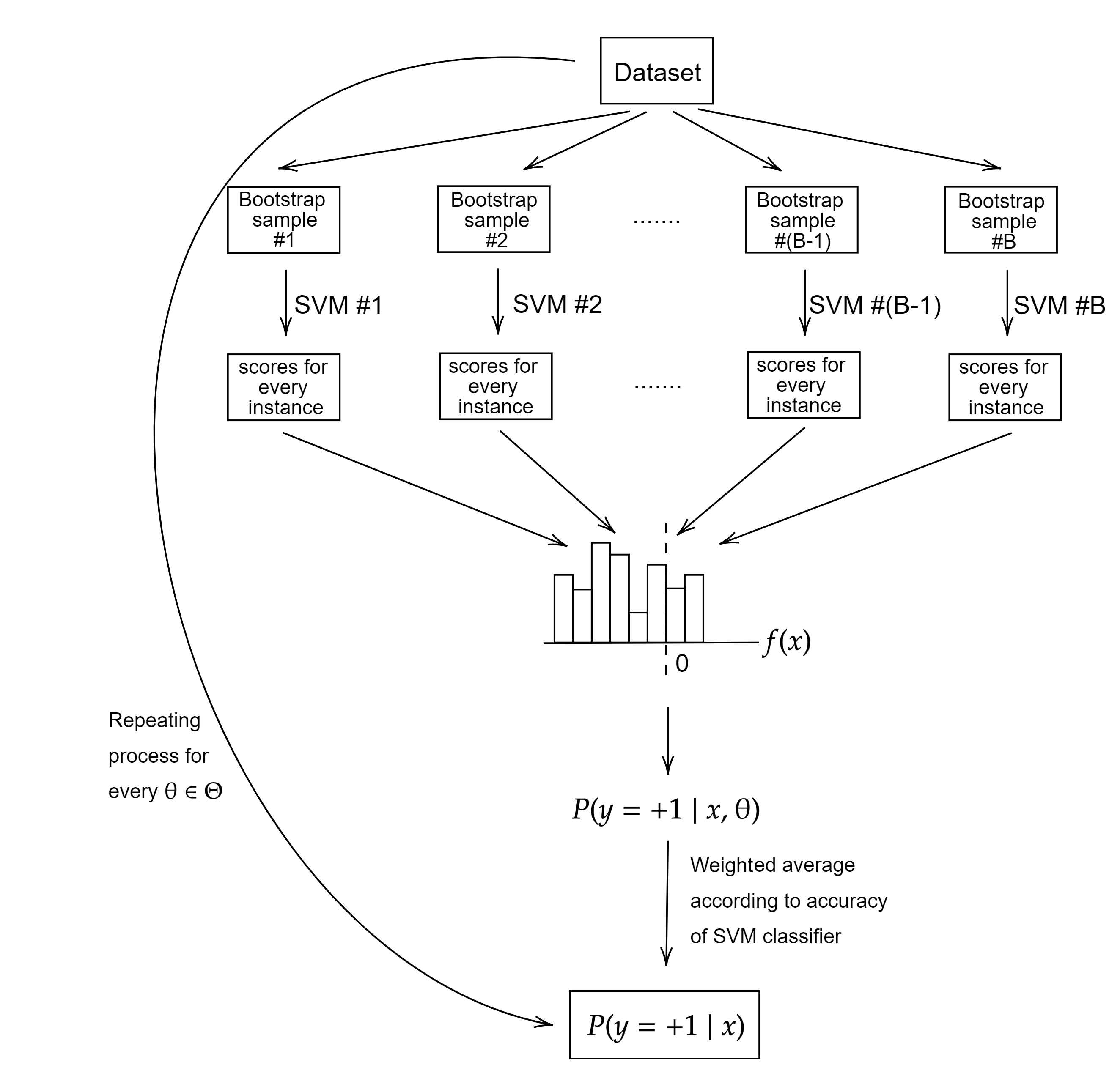}
  \caption{Flowchart of the Bootstrap-based methodology for estimating $P(y=+1 \mid x)$.}\label{fig:bootstrapflow}
\end{figure}

\begin{figure}
\centering
\includegraphics[width=12cm]{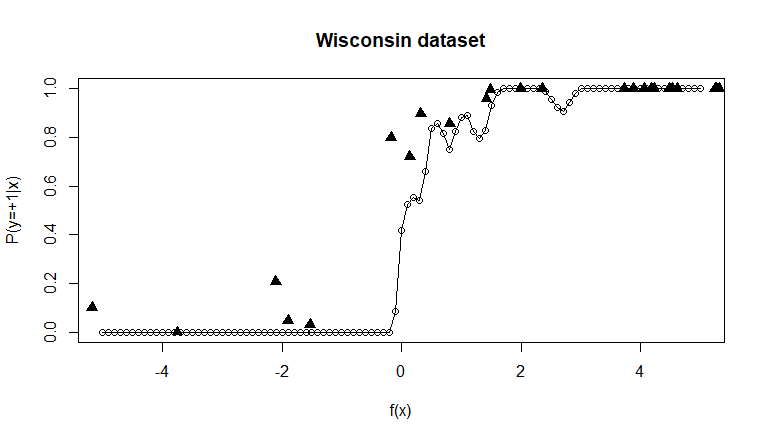}
\caption{Solid line with dots represents the empirical class probabilities with \textit{EBB}. Triangles show the probabilities obtained under the novel method in an external validation sample.}\label{fig:wisconsin_scores_2}
\end{figure}

In particular, the methodology that we propose, the Ensembled Bootstrap-Based (\textit{EBB}), is carried out as follows. First, we generate, from our original {training dataset $T$}, a total of $B$ bootstrap samples. For each of these samples, using a fixed parameter value $\theta ~\in~ \Theta$, we build a SVM. Hence, score values for {every} instance in the original dataset $\Omega$ can be easily calculated. We will obtain this way a total of $B$ score values for every instance in $\Omega$. Therefore, for a given instance with attribute vector $x$, its class probability $P(y = +1 ~ | ~ x, \theta)$ (or $P(y = -1 ~ | ~ x, \theta)$) will be calculated as the proportion of positive (respectively, negative) scores obtained for such instance. To eventually obtain the class probability $P(y = +1 ~ | ~ x)$ we will take into account the reliability $\rho_{\theta}$ of the different parameters $\theta ~\in ~\Theta$. In order to define such a reliability index $\rho_{\theta}$, we can consider a classical performance measure such as the accuracy (or the AUC, the G-Mean, etc), that we will denote as $acc_{\theta,b}$, for a given parameter $\theta \in \Theta$ and the $b$-th bootstrap sample. We can estimate the overall accuracy $\overline{acc}_{\theta}$ using the parameter $\theta \in \Theta$, as the average value of the coefficients $acc_{\theta,b}$ in the different bootstrap samples, namely
\begin{equation}\label{eq:AverageAccuracy}
\overline{acc}_{\theta} = \dfrac{\sum\limits_{b=1}^B acc_{\theta,b}}{B}.
\end{equation}
As commented previously, our purpose is to build an ensemble classifier using the information obtained during the parameter tuning process. Although all the obtained classifiers could be used in this ensemble, we can also discard the worse ones by defining the set $J = \{ j : \overline{acc}_{\theta_j} \geq \max_l \overline{acc}_{\theta_l} - \varepsilon\}$, where $\varepsilon > 0$ is a fixed parameter. Finally, we can determine the reliability $\rho_{\theta}$ as

\begin{equation}\label{eq:Averagewei}
    \rho_{\theta_j} =
    \dfrac{\overline{acc}_{\theta_j}}{\sum\limits_{l \in J} \overline{acc}_{\theta_l}},
\end{equation}
$\text{if } ~ j \in J, \text{ and } \rho_{\theta_j} = 0$ otherwise.
Finally, \textit{EBB} estimates the posterior class probability as
\begin{equation}\label{eq:Average}
P(y = +1 ~ | ~ x) = \sum_{j \in J} \rho_{\theta_j} P(y = +1 ~ | ~ x,\theta_j).
\end{equation}
This procedure is depicted in Figure~\ref{fig:bootstrapflow}. Moreover, in Figure~\ref{fig:wisconsin_scores_2} we have created analogous plots as the ones in Figure~\ref{fig:PlattProced}, but using our \textit{EBB} methodology. \textcolor{black}{The reader can observe that our approach, which is the one depicted in Figure~\ref{fig:wisconsin_scores_2}, clearly fits better the \texttt{wisconsin} dataset by comparing it with Figure~\ref{fig:PlattProced}, where Platt's approach is used.}

\textcolor{black}{%As commented before, the bootstrap sample allows us to obtain point estimates for the posterior class probabilities $P(y=+1 \mid x)$. From the classic confidence interval for a proportion, it is straightforward to construct a $(1-\alpha)\%$ confidence interval for $\hat{p} = P(y=+1 \mid x)$ as
}

The novel methodology will be illustrated in Section \ref{subsec:LinearResults} where, in addition, some comparisons with respect to benchmark approaches will also be presented.

\subsection{Class probabilities estimates after controlling the sensitivity of the classifier}\label{subsec:CSSM}

In the previous section, the \textit{EBB} approach for estimating posterior class probabilities \textcolor{black}{$P(y=+1\mid x)$ and $P(y=-1\mid x)$} has been described. In this section, we deal with the issue of improving the sensitivity of the classifier (or TPR) which, as commented in Section~\ref{Intro}, may be a problem of interest, among others, \textcolor{black}{in business, social sciences or biomedical contexts. To do this, we propose two different approaches, the \textit{ Cost-sensitive Ensembled Bootstrap-Based 1} (\textit{CEBB$_1$}) and the \textit{Cost-sensitive Ensembled Bootstrap-Based 2} (\textit{CEBB$_2$}), which are discussed in what follows and are empirically analyzed in Section~\ref{subsec:CostSensit}.}

\begin{figure}
\centering     %%% not \center
\subfigure[]{\label{fig:6a}\includegraphics[width=60mm]{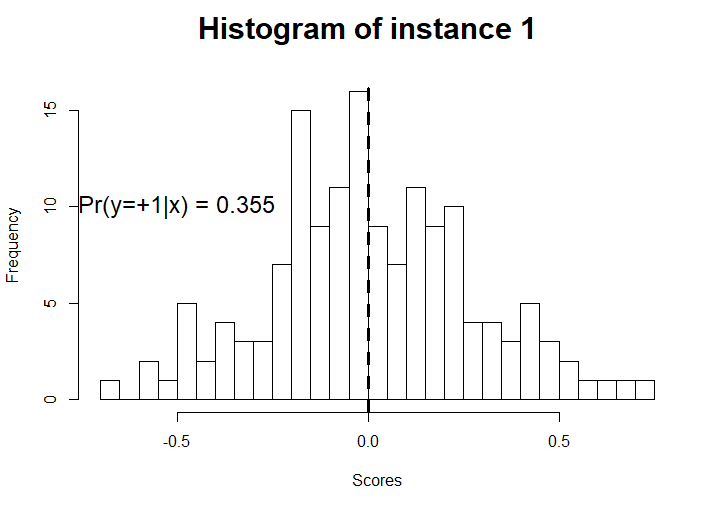}
\includegraphics[width=60mm]{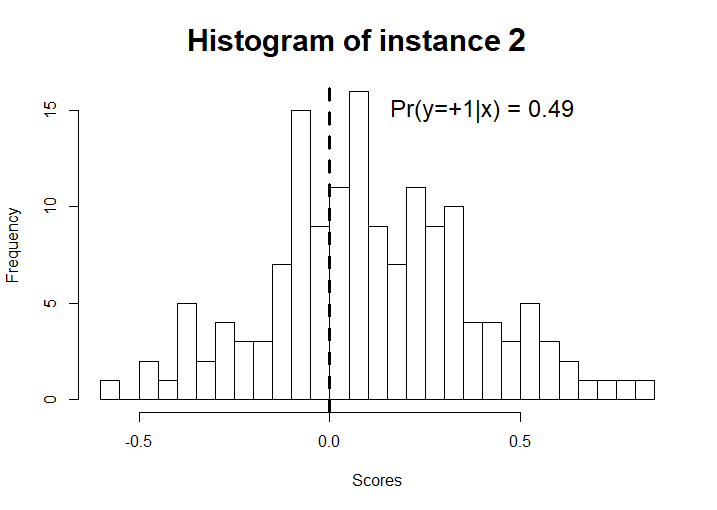}}
\newline
\centering
\subfigure[]{\label{fig:6b}\includegraphics[width=60mm]{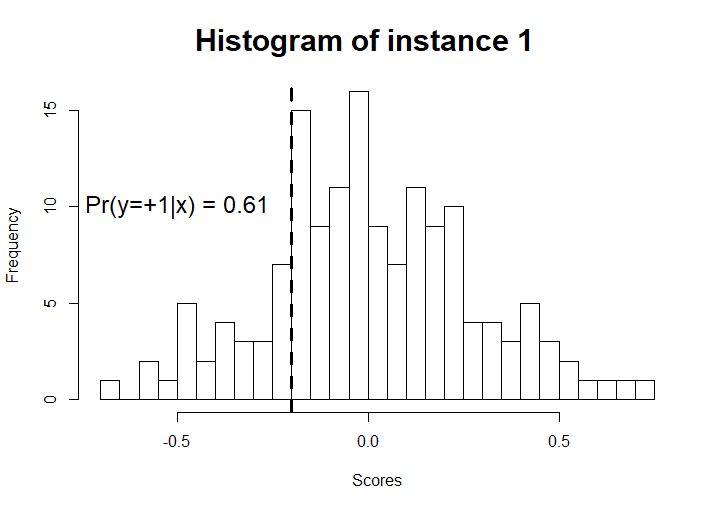}
\includegraphics[width=60mm]{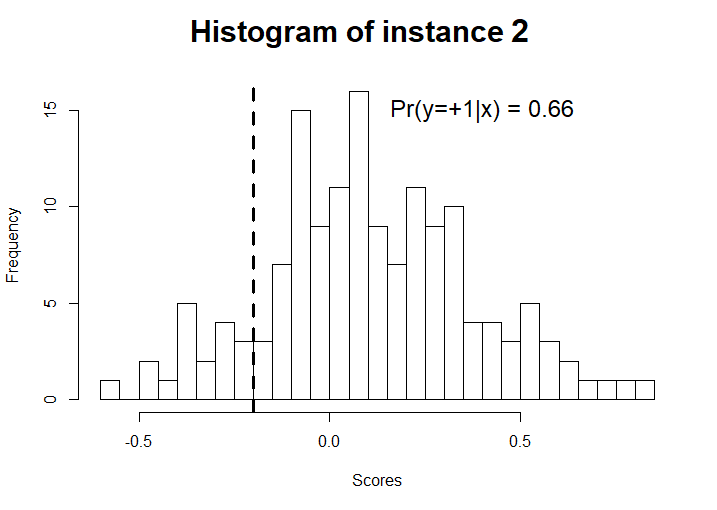}}
\caption{Control over the probabilities estimation. In Subfigures~\ref{fig:6a} we can observe the original estimated probabilities, whereas in Subfigures~\ref{fig:6b} the new cost-sensitive probabilities for \ref{fig:6a}, obtained by moving the threshold, are depicted.}
\label{fig:6}
\end{figure}

Method \textit{CEBB$_1$} is based on the fact that the sensitivity measure can be controlled by the posterior class probabilities, as is explained next. In the previous section, the posterior negative class probabilities have been estimated taking into account the proportion of negative scores. However, if instead of $0$, we consider a different threshold (say a value $-a$, with $a$ positive), then the estimates \textcolor{black}{for} $P(y = -1 \mid x, \theta)$ decrease (that is, the posterior positive class probabilities  \textcolor{black}{$P(y = +1 \mid x, \theta)$} increase). This is illustrated by two examples in Figure~\ref{fig:6}, Figures~\ref{fig:6a} and \ref{fig:6b}, which represent the histograms of the scores for two different individuals of the \texttt{churn} dataset randomly chosen. Figure~\ref{fig:6a} shows the  posterior positive class probability estimates using \textit{EBB}, that is, where the value $0$ is used as a threshold to classify in the positive or the negative class. In Figure~\ref{fig:6b}, the threshold value has been moved to the left and, as a consequence, the resulting estimates have increased. Note from Figure~\ref{fig:6b} that, with this approach, the probability of an instance to belong to the positive class may change from below to above 0.5. In practice, in order to obtain a desired posterior positive class probability estimate, the threshold is moved until a certain proportion of the instances of the positive class are correctly classified.

Method \textit{CEBB$_2$} works similar to \textit{CEBB$_1$}, but, instead of changing the threshold for the scores, we consider a different classifier. Specifically, we propose to use a novel version of the SVM, the so-called Constrained SVM (CSVM), which has been particularly designed to obtain cost-sensitive results, see \cite{Sandra2}. Without going into much detail, the CSVM formulation is obtained by solving a convex quadratic optimization problem with linear constraints and some integer variables:

%Using the bootstrap procedure allows us not only to obtain probabilities for the obtained classification but also to control those probabilities. Controlling the probabilities refers for example to, if we know that given a set of instances belong to a certain class, but the probability of belonging to such class is low, modify the classifier or the probabilities estimation so that such probabilities are increased but the other that were correct keep approximately the same.

%Here, we propose two different options to control the classification probabilities. The first one is purely based on the bootstrap procedure, but we change the way we calculate the probabilities based on it, as will be explained in Section~\ref{sssec:ChanProb}. Basically, this first proposal consists on changing the threshold score from which an instance is considered as positive or negative. Originally, this threshold is set in the score $0$; however, we will move such threshold until a given proportion of instances are correctly classified. The second one is based on changing the classifier by that cost-sensitive one presented in \cite{Sandra2,Sandra1},
\begin{equation}
\label{eq:csvm}
\begin{array}{lll}
  \min_{\omega, \beta, \xi, \zeta} & \omega^\top \omega+ C\sum_{i \in I} \xi_i& \\
  s.t. & y_i(\omega^\top {x}_i + \beta) \geq 1 - \xi_i,& i \in I \\
   & 0 \leq \xi_i \leq M(1-\zeta_i)& i \in I\\
   & \mu(\zeta)_\ell \geq \lambda_\ell & \ell \in L\\
   & \zeta_i \in \{0,1\} & i \in I.
\end{array}
\end{equation}

Problem (\ref{eq:csvm}) is simply the formulation for the standard SVM with linear kernel, to which performance constraints have been added: $\mu(\zeta)_\ell \geq \lambda_\ell$, where $\mu(\zeta)_\ell$ are different performance measures, forced to take values above thresholds $\lambda_\ell$, and $\zeta_i$ are new binary variables that check whether record $i$ is counted as correctly classified, and $M$ is a large number. We refer the reader to the original reference (\cite{Sandra2}) for a more detailed description of the cost-sensitive classifier. The important message to be kept here is that solving (\ref{eq:csvm}) with a standard software package for different values of the parameters $\lambda_\ell$ yields classifiers with different trade-off between sensitivity and specificity.

Both methods (\textit{CEBB$_1$} and \textit{CEBB$_2$}) will be illustrated through numerical examples in Section \ref{subsec:CostSensit}{, where we will be able to better see and explain their differences and similarities, as well as the advantages and disadvantages of each.}

\section{Experimental results}\label{sec:ER}

In this section we illustrate the performance of  the novel $EBB$ method for computing posterior class probability estimates as described in  Section~\ref{subsec:BootsProp}. The results will be compared to those of benchmark approaches by \cite{platt1999probabilistic} {(which can be found implemented in recognized programming languages such as Python or R, as previously commented)}, \cite{sollich2002bayesian} and \cite{tao2005posterior}. For that, a variety of datasets with different properties concerning size (in the number of instances and/or variables) and imbalancedness shall be analyzed. Moreover, we test the methods described in Section~\ref{subsec:CSSM} to control the posterior positive class probability. {The experiments were run on a computer with an Intel\textregistered ~ Core\texttrademark ~ i7-6700 processor at 3.4 GHz using 16 GB of RAM, running Windows 10 Home. All the optimization problems have been solved using Python 3.9 interface (\cite{van1995python}) with Gurobi 9.1.2 solver (\cite{gurobi}).} This section is organized as follows. In Section~\ref{subsec:ExpDesc} we describe how the different experiments have been implemented. Then, in Section~\ref{subsec:DataExp}, we present a brief description of the different datasets we have used. Section~\ref{subsec:LinearResults} shows the performance of the novel approach \textit{EBB} in comparison to benchmark methodologies \textcolor{black}{to obtain point estimates of the probabilities}. Finally, in Section~\ref{subsec:CostSensit} we apply both \textit{CEBB$_1$} and \textit{CEBB$_2$} to improve the posterior probability of the class of interest.

\subsection{Experiment description}\label{subsec:ExpDesc}

Now we explain how our procedure will be implemented. The pseudocode of our method can be found in Algorithm~\ref{algBoot}.

\RestyleAlgo{boxruled}
\begin{algorithm}
  \caption{Pseudocode for the Bootstrap SVM}\label{algBoot}
  \begin{algorithmic}[1]
  \STATE Inputs: $T$ (training sample), $B$ (number of bootstrap replicates),\\ $\Theta$ (grid of parameters to tune SVM classifier)
  \STATE Outputs: $B\times|\Theta|$ SVM models and their accuracies.
    \FOR{each value $\theta$ in $\Theta$}
    \FOR{$b$ in $1,2,\ldots,B$}
      \STATE Create a bootstrap sample $T^*_b$ from $T$.
      \STATE Build a SVM using $T_b^*$. Validate over $V_b^* = T\setminus T_b^*$.
      \STATE Obtain for each sample $V_b^*$ its performance $acc_{\theta,b}$.
    \ENDFOR{}
    \STATE Calculate the average accuracy $\overline{acc}_\theta$ for SVM with parameter $\theta$\\ as in Eq.~(\ref{eq:AverageAccuracy}).
    \ENDFOR{}
  \end{algorithmic}
\end{algorithm}

The explanation of the steps provided by Algorithm~\ref{algBoot} is given next.
First, consider {our complete training dataset $T$} composed of $m$ instances, $n$ variables, and 2 classes ($+1$ or $-1$). {For this dataset, we will assume that the class label information is known, in order to train the SVM and estimate the accuracy of the classifier}. As it is usual in the SVM implementation, a grid $\Theta$ for the parameters needs to be \textcolor{black}{set.}
%, in this case, 11 values of $C$ will be tuned: $\{C_1,...,C_{11}\} = \{2^{-5},2^{-4},...,2^{4},2^{5}\}$.

%Then, a matrix $PX$ \textcolor{black}{with as many rows as the number of instances in the outer sample ($m-mtr$) and as many columns as the number of values  $\theta \in \Theta$ to be used is built. This matrix contains the proportions of negative scores %(and illustrated in Table~\ref{tab:PX})
 % and is generated through the algorithm as follows.}

Given a fixed value $\theta \in \Theta$, the next step is to compute $B$ bootstrap samples (Step 5) of size $m$ (denoted as $T_b^*$, $b=1,\ldots,B$) from the training sample $T$. Note that, according to Efron and Tibshirani (1997), {only around a $63.2\%$ of the elements in the original dataset are collected in a bootstrap sample}. Then, a SVM classification is built using each bootstrap sample (Step 6) {and validated over the out of bag sample, i.e., the set of instances that are in $T$ but not in the considered bootstrap sample (we will denote them as $V^*_b, ~b = 1,\ldots,B$){, with irregular size, depending on the number of unique instances in $T^*_b$}.}

{The outputs of the Algorithm are the total of $B\times |\Theta|$ SVM models and their accuracies. They will be used to estimate, for any given instance in $\Omega$, the posterior negative class probabilities $P(y_{i}=-1\mid x)$ as a weighted average from the reliability indexes as in (\ref{eq:Averagewei}) and calculated as in (\ref{eq:Average}).
}

We will present the specific details for \textit{EBB}, \textit{CEBB$_1$} and \textit{CEBB$_2$} in what follows.
When running the {standard} SVM (for \textit{EBB} and \textit{CEBB$_1$}) and the constrained SVM (for \textit{CEBB$_2$}) in (\ref{eq:svm}) or (\ref{eq:csvm}), the linear kernel versions were considered. {As previously commented, all} the experiments have been carried out using the solver Gurobi (\cite{gurobi}) and its Python language interface (\cite{pthn}). No timelimit was imposed when solving Problem~(\ref{eq:svm}), whereas a 300-second timelimit was set when solving (\ref{eq:csvm}). Also, for the latter problem, $M$ was equal to 1000 (see, \cite{Sandra2} for more details).

Our experiments are separated in two parts. First, we compare our methodology \textit{EBB} with benchmark approaches presented in Section~\ref{Intro}. Second, we demonstrate how we can control the posterior probability errors using \textit{CEBB$_1$} and \textit{CEBB$_2$}.

\textcolor{black}{For the benchmark methodologies{, since their resulting probabilities are not dependent on the value $k$}, a standard $k$-fold CV will be carried out. Here, we will use $k=10$ external folds (in order to estimate the performance measure by the average over these folds) and $k=10$ internal folds (for obtaining the best parameter $\theta \in \Theta$). On the other side, in our methodologies \textit{EBB}, \textit{CEBB$_1$} and \textit{CEBB$_2$}, the bootstrap approach will be used {in order to avoid its previously commented dependence on $k$ and degenerate behaviour}. The number of bootstrap samples $B$ will be set equal to $500$ and each bootstrap training sample has the same size as the original training sample. Note that we cope with the imbalancedness, if present, though one could have performed under or oversampling in the majority or the minority class, respectively, in a preprocessing phase. Since we are using the linear kernel, we only have the parameter $\theta = C$. The grid $\Theta$ of values selected in our experiments is $\{2^{-5},2^{-4},...,2^{4},2^{5}\}$.}

\subsection{Datasets description}\label{subsec:DataExp}

The performance of the different methodologies presented in this paper is illustrated using fourteen real-life datasets: \texttt{absenteeism} (Absenteeism at work Data Set), \texttt{adult} (Adult), \texttt{australian} (Statlog (Australian Credit Approval) Data Set), \texttt{banknote} (banknote authentication), \texttt{careval} (Car Evaluation Data Set), \texttt{cervical-cancer} (Cervical cancer (Risk Factors)), \texttt{churn} (Customer churn), \texttt{german} (German Credit Data), \texttt{heart} (Heart Disease), \texttt{housing} (The Boston Housing Dataset), \texttt{leukemia} (Leukemia), \texttt{productivity} (Productivity Prediction of Garment Employees Data Set), \texttt{SRBCT} (Small Round Blue Cell Tumor) and  \texttt{wisconsin} (Breast Cancer Wisconsin (Diagnostic)).

%\texttt{cancer-colon} (Colon Cancer),
\texttt{SRBCT} dataset can be obtained from the R package \texttt{plsgenomics} (\cite{boulesteix2011plsgenomics}) and \texttt{leukemia} from \cite{golub1999molecular}. On the other hand, %\texttt{cancer-colon} is available at the Kent Ridge Biomedical Data Repository (\cite{ridge2002kent}),
\texttt{housing} is taken from \cite{HARRISON197881} and \texttt{churn} from \cite{keramati2011churn}. The other eleven datasets are obtained from the UCI Repository, (\cite{Dua2017}). Dataset \texttt{cervical-cancer} has been split into two different datasets since it contains $4$ different variables of class. We show two of them as an illustration.
Table~\ref{tab:data} contains relevant information of the previous datasets. {In the first column, we can find the name of the different datasets.} In the second{, third  and fourth} columns, the sample sizes of the validation {($|V|$)}{, outer ($|T|$)/inner ($|T^*_b|$) training,} and the complete datasets {($|\Omega|$, where $\Omega$ = $T\cup V$ here)} are shown, respectively. The fifth column contains the number of original variables or attributes {($|A|$)} in the dataset. Finally, the last column collects the number {($|\Omega_{+}|$)} and percentage {($\%$)} of positive instances \textcolor{black}{in the complete dataset}.
\begin{table}[h!]

\centering \small

\begin{tabular}{lllllr}

     \hline
  \textit{Name} & {\textit{$|V|$}} & {\textit{$|T| = |T^*_b|$}}  & \textit{$|\Omega|$} & \textit{$|A|$} & \textit{$|\Omega_+|$ \textit{(\%)}} \\
  \hline
    \texttt{absenteeism} & 74 & 665 & 739 & 20 & 272 (36.81\%)\\
    \texttt{adult} & 3256 & 29305 & 32561 & 14 & 7841 (24.08\%) \\
    {\texttt{australian}} & 69 & 621 & 690 & 14 & 307 (44.49\%) \\
    \texttt{banknote} & 137 & 1235 & 1372 & 5 & 610 (44.46\%) \\
    %\texttt{cancer-colon} & 6 & 62 & 2000 & 22 (35.48\%) \\
    {\texttt{careval}} & 173 & 1555 & 1728 & 6 & 518 (29.98\%)  \\
    \texttt{cervical-cancer-1} & 86 & 772 & 858 & 36 & 35 (4.08\%)\\
    \texttt{cervical-cancer-2} & 86 & 772 & 858 & 36 & 74 (8.62\%) \\
    %\texttt{cervical-cancer-3} & 86 & 858 & 36 & 44 (5.13\%) \\
    %\texttt{cervical-cancer-4} & 86 & 858 & 36 & 55 (6.41\%)\\
    \texttt{churn} & 315 & 2835 & 3150 & 11 & 495 (15.71\%) \\
    %\texttt{diabetes} & 77 & 768 & 20 & 500 (65.1\%) \\
    %\textcolor{black}{\texttt{divorce}} & 17 & 170 & 54 & 84 (49.41\%) \\
    {\texttt{german}} & 100 & 900 & 1000 & 20 & 300 (30\%) \\
    \texttt{heart} & 72 & 648 & 720 & 75 & 362 (50.28\%)\\
    \texttt{housing} & 51 & 455 & 506 & 13 & 256 (50.59\%) \\
    \texttt{leukemia} & 7 &65 & 72 & 7128 & 25 (34.72\%)  \\
    \texttt{productivity} & 120 & 1076 & 1196 & 14 & 474 (39.63\%) \\
    \texttt{SRBCT} & 8 &75 &83 & 1022 & 29 (34.94\%) \\
    \texttt{wisconsin} & 57 & 512& 569 & 30 & 212 (37.26\%) \\
    \hline

 \end{tabular}

\caption{Datasets}

\label{tab:data}

\end{table}

\textcolor{black}{In a pre-processing step, the categorical variables were transformed into a set of dummy variables using a one hot encoding. In addition, those datasets with three classes or more were converted into two-class datasets by giving negative label to the largest class and positive labels to the remaining records. In the case of missing values, they were replaced by the median in the case of numerical variables and by the mode in the case of categorical ones. Standardization of the data to have each numerical variable coming from a distribution with mean $0$ and unit variance has been consider in each fold (for both the $k$-fold CV and the bootstrap), performing it first over the training {($T$)} data and then using the obtained average and standard deviation to standardize the validation {($V$)} one.}

%Now, we will explore first how to calculate the probabilities and then, how to control them.

%Now, we are in conditions of explaining how to calculate the class probabilities using the SVM as the basis classifier. In addition, we will use the properties of the bootstrap to include not only a punctual estimation but also a confidence interval for the probabilities. In the following sections, we will explain how the experiments have been carried out. All our experiments have been solved using the solver Gurobi (\cite{gurobi}) and its Python language interface (\cite{pthn}). No timelimit was imposed when solving Problem~(\ref{eq:svm}), whereas 300 seconds was set when solving (\ref{eq:csvm}). Also, for this last problem, $M$ was equal to 1000. Now, we will explore first how to calculate the probabilities and then, how to control them.

\subsection{Performance of the ensembled bootstrap-based approach (EBB)}\label{subsec:LinearResults}

%SVM posterior class probabilities based on Bootstrap}\label{subsec:BootsProp}

In this section we obtain point estimates of the posterior class probabilities according to the ensembled bootstrap-based (\textit{EBB}) novel method described in Section \ref{subsec:BootsProp} and compare the results with those obtained by the benchmark approaches by \cite{platt1999probabilistic}, \cite{sollich2002bayesian} and \cite{tao2005posterior} commented in Section 1. The obtained results are summarized in Table~\ref{tab:ResultsLinear}, whose columns contain the mean squared errors \textcolor{black}{(MSE)} values obtained when the deterministic class membership is compared with its probabilistic counterpart for all the methods. Particularly, if we have $p_i = P(y_i = +1)$, here assumed to be $p_i \in \{0,1 \}$ (since we are given the class labels, we know the actual value $p_i$) and its estimate $\hat{p}_i$, we calculate the results in Table~\ref{tab:ResultsLinear} as
\begin{equation}\label{eq:MSE}
MSE = \dfrac{\sum\limits_{i \in V}^n(p_i - \hat{p}_i)^2}{|V|}.
\end{equation}
In the first column of Table~\ref{tab:ResultsLinear}, we can observe the name of the dataset. In the second and third ones, the MSE as in (\ref{eq:MSE}) when our method using just the best value of $\theta$ and all the values of $\theta$ are used, respectively. In the fourth, fifth and sixth columns we can see the results when using the methods in \cite{sollich2002bayesian}, \cite{platt1999probabilistic} and \cite{tao2005posterior}, respectively.

Note that, according to \cite{tao2005posterior}, a value for the parameter $r$ needs to be selected. In this case, we tested the results for four different choices of $r$ ($0$, $\sqrt{10}$, $\sqrt{20}$, $\sqrt{30}$). The best results have been highlighted in bold style. {The same type of results, but using as performance method the AUC (area under the curve) instead of the MSE are presented in the Supplementary Material of this manuscript.}

\begin{comment}
\begin{table}[h!]

\centering \small

\begin{tabular}{lllll}

     \hline
  \text{Dataset} & \text{Best} \theta & \textit{\text{EBB}} & \textit{Sollich} & \textit{\text{Platt}} &\textit{\text{Tao et al.}} \\
  & & & & ($r=0$, $\sqrt{10}$, $\sqrt{20}$, $\sqrt{30}$)\\
  \hline
  \texttt{wisconsin} & \textbf{0.003}  &  0.064 &  0.021  & 0.028, 0.019, 0.034, 0.055 \\
  \texttt{cancer-colon} & \textbf{0.201}  & 0.241  & 0.252 & 0.208, 0.208, 0.208, 0.208  \\
  \texttt{diabetes} & 0.192 & 0.190  & \textbf{0.157} & 0.229, 0.233, 0.234, 0.234 \\
  \texttt{leukemia} & \textbf{0}  & 0.239  &  0.01 &  0.029, 0.029, 0.029, 0.029  \\
  \texttt{SRBCT} &  0.01 & 0.237  & 0.011 & \textbf{0}, \textbf{0}, \textbf{0}, \textbf{0} \\
  \texttt{heart} & 0.143  & 0.158  & \textbf{0.121} & 0.15, 0.118, 0.207, 0.218 \\
  \texttt{adult} &    0.144    &    0.128    & \textbf{0.068} & 0.148, 0.078, 0.142, 0.167\\
  \texttt{divorce} & \textbf{0} & 0.189 & 0.021 & 0.024, 0.024, 0.024, 0.024\\
  \texttt{german} & 0.197 & 0.229 & \textbf{0.168} & 0.256, 0.256, 0.256, 0.247\\
  \texttt{cervical-cancer} & \textbf{0.013} & 0.187 & 0.028 & 0.039, 0.039, 0.032, 0.035 \\
  \texttt{banknote} & 0.017 & \textbf{0.008} & \textbf{0.008} & 0.009, 0.145, 0.221, 0.238\\
    \hline

 \end{tabular}

\caption{Out-of-sample mean squared errors (MSE) obtained when predicting the posterior class probabilities in a linear SVM.}
\label{tab:ResultsLinear}

\end{table}
\end{comment}

\begin{table}[ht]
\centering \small
\begin{tabular}{llllll}
     \hline
  \textit{Dataset} &  \textit{EBB (\textit{Best} $\theta$)} & \textit{\text{EBB}} & \textit{Sollich} & \textit{\text{Platt}} & \textit{\text{Tao et al.}} \\
  & & & & & ($r=0$, $\sqrt{10}$, $\sqrt{20}$, $\sqrt{30}$)\\
  \hline
    \texttt{absenteeism} & 0.176 & \textbf{0.133} & 0.16 & \textbf{0.133} & 0.176, 0.176, 0.176, 0.176 \\
    \texttt{adult} & 0.163 & 0.158 & 0.232 & \textbf{0.105}  & 0.151, 0.146, 0.128, 0.13\\
    \texttt{australian} & 0.217 & 0.173 & 0.223 & \textbf{0.121} & 0.151, 0.149, 0.135, 0.133\\
    \texttt{banknote} & \textbf{0.007} & 0.011 & 0.008 & 0.09 & 0.008, 0.145, 0.221, 0.238\\
    %\texttt{cancer-colon} & 0.177 & 0.238 & \textbf{0.156} &   0.174, 0.174, 0.174, 0.174\\
    \texttt{careval} & 0.055 & 0.052 & 0.074 & \textbf{0.04} & 0.047, 0.047, 0.043, 0.093\\
    \texttt{cervical-cancer-1} & \textbf{0.012} & 0.013 & 0.234 & 0.04 & 0.045, 0.045, 0.045, 0.045\\
    \texttt{cervical-cancer-2} & 0.105 & \textbf{0.075} & 0.199 & 0.08 & 0.103, 0.103, 0.103, 0.103 \\
    %\texttt{cervical-cancer-3} & 0.058 &  0.236 & \textbf{0.049} & 0.055, 0.055, 0.055, 0.055\\
    %\texttt{cervical-cancer-4} & 0.075 & 0.175 & \textbf{0.061} & 0.075, 0.075, 0.075, 0.075\\
    \texttt{churn} & 0.108 & 0.108 & 0.227 & \textbf{0.093} & 0.104, 0.104, 0.101, 0.128\\
    %\texttt{diabetes} & 0.196 & 0.219 & \textbf{0.159} & 0.224, 0.161, 0.210, 0.228\\
    %\texttt{divorce} & 0.059 & 0.124 & \textbf{0.016} & 0.024, 0.024, 0.024, 0.024\\
    \texttt{german} & 0.24 & 0.203 & 0.203 & \textbf{0.163} & 0.235, 0.224, 0.187, 0.182\\
    \texttt{heart} & 0.194 & 0.171 & 0.217 & 0.124 & 0.157, 0.119, \textbf{0.113}, 0.168 \\
    \texttt{housing} & 0.118 & \textbf{0.078} & 0.142 & 0.097 & 0.152, 0.118, 0.141, 0.163\\
    \texttt{leukemia} & 0 & \textbf{0} & 0.238 & 0.019 & 0.014, 0.014, 0.014, 0.014\\
    \texttt{productivity} & 0.286 & 0.212 & 0.238 & \textbf{0.190} & 0.242, 0.239, 0.262, 0.259 \\
    \texttt{SRBCT} & 0.125 & 0.039 & 0.237 & 0.006 &  \textbf{0}, \textbf{0}, \textbf{0}, \textbf{0}\\
    \texttt{wisconsin} & 0.035 & \textbf{0.018} & 0.094 & 0.034 & 0.028, 0.019, 0.037, 0.055\\
    \hline
 \end{tabular}
\caption{Out-of-sample mean squared errors (MSE) obtained when predicting the posterior class probabilities in a linear SVM.}
\label{tab:ResultsLinear}

\end{table}

\textcolor{black}{
{It can be seen from Table~\ref{tab:ResultsLinear} that our methodology \textit{EBB} is the one performing best for \texttt{absenteeism}, \texttt{banknote} (in this case, using just the best $\theta$, which is a degenerate case of \textit{EBB} with $\varepsilon=0$), \texttt{cervical-cancer-1} (using again the best $\theta$), \texttt{cervical-cancer-2}, \texttt{housing}, \texttt{leukemia} and \texttt{wisconsin}, obtaining the lowest values of MSE. Additionally, the method proposed by \cite{platt1999probabilistic} obtains the lowest MSE in \texttt{absenteeism}, \texttt{adult}, \texttt{australian}, \texttt{careval}, \texttt{churn}, \texttt{german} and \texttt{productivity}. Finally, with the method of \cite{tao2005posterior}, the lowest MSE is obtained in \texttt{heart}, and also a zero MSE for \texttt{SRBCT} is achieved. On the other hand, the method proposed by \cite{sollich2002bayesian} performs poorly in all cases}. In conclusion, we have built a method for obtaining point estimates that is comparable in terms of performance to benchmark approaches, outperforming them in some datasets. Furthermore, we have empirically demonstrated that, in general, \textit{EBB} method with a weighted average of different SVM with distinct parameters $\theta \in \Theta$ (that is, when ensembled method) works better than when only using the best value $\theta \in \Theta$.}

{At this point it is of interest
to analyze the performance of the method when the datasets are unbalanced. With this aim, we have considered from Table~\ref{tab:data} the six most unbalanced samples, which are the two ``cervical-cancer'' datasets, followed by ``churn'', ``adult'', ``careval'' and ``german''. From Table~\ref{tab:ResultsLinear}, we can see how in the two most unbalanced datasets (cervical-cancer 1 and 2) our methods outperform the others, while in the other 4 unbalanced datasets, Platt is better (in the particular case of ``adult'', we know in advance that it adheres quite well to the sigmoid function proposed by Platt). The next section illustrates how the cost-sensitivity of our method allows from improving the results regarding these four unbalanced datasets.}

\subsection{Results when the posterior class probabilities are controlled (\textit{CEBB$_1$} and \textit{CEBB$_2$})}\label{subsec:CostSensit}

In this section we apply the {cost-sensitive} methodologies described in Section~\ref{subsec:CSSM} (\textit{CEBB$_1$} and \textit{CEBB$_2$}) {in order to control $P(y=+1\mid x)$. In particular, Table~\ref{fig:CSv11prueba} shows the MSE results for the positive class that are obtained using both \textit{CEBB$_1$} and \textit{CEBB$_2$}. Here, the class of interest to be controlled is assumed to be the positive one\textcolor{black}{, that is, we aim to control the true positive rate (TPR).} Table~\ref{fig:CSv11prueba}
shows the MSE as defined in (\ref{eq:MSE}) when considering only the actual positive instances.}

From Table~\ref{fig:CSv11prueba}, we can see how as the threshold for obtaining a given proportion of the instances in the correct class is increased, the MSE becomes lower, as expected. {In fact, there are some datasets (\texttt{banknote}, \texttt{careval}, \texttt{heart}, \texttt{housing}, \texttt{SRBCT} and \texttt{wisconsin}), for which the obtained MSEs are very close to $0$.
%%%%%%%%%%%%%%%%%%% PRUEBA %%%%%%%%%%%%%%%
\begin{table}[h!]
\Rotatebox{0}{
\centering
\begin{tabular}{|cc|ccccccc|}
  \hline
  \multicolumn{2}{|c|}{\text{Dataset (Method) $\backslash$ TPR imposed}} & 0 &  0.5 & 0.6 & 0.7 & 0.8  & 0.9 & 1\\
  \hline
        %\texttt{adult} & 0.163 & 0.158 \\
    \multirow{2}{*}{\texttt{absentism}} & ($CEBB_1$) & 0.266 & 0.266 & 0.266 & 0.266 & 0.266  & 0.266 &  0.266\\
    & ($CEBB_2$) & 0.266 &  0.266 & 0.266 & 0.266 & 0.254  & 0.237 &  0.221\\
    \hline
    \multirow{2}{*}{\texttt{adult}} & ($CEBB_1$) & 0.476 & 0.476 & 0.476 & 0.476 & 0.476 & 0.476 & 0.476 \\
    & ($CEBB_2$) & 0.476 & 0.476 & 0.112 & 0.003 & 0.000  & 0.000 & 0.219\\
    \hline
    \multirow{2}{*}{\texttt{australian}} & ($CEBB_1$)  & 0.071 & 0.071 & 0.071 & 0.071 & 0.071  & 0.071 &  0.070 \\
    & ($CEBB_2$) & 0.071 & 0.071 & 0.071 & 0.071 & 0.071  & 0.056 & 0.091\\
    \hline
    \multirow{2}{*}{\texttt{banknote}} & ($CEBB_1$)  & 0 & 0 & 0 & 0 & 0 & 0  &  0 \\
    & ($CEBB_2$) & 0 & 0 & 0 & 0 & 0 &  0 & 0 \\
    \hline
    %\texttt{cancer-colon} & 0.167 & 0.177 \\
    \multirow{2}{*}{\texttt{careval}} & ($CEBB_1$)   & 0.044 & 0.044 & 0.044 & 0.044 & 0.044  & 0.044  & 0.038 \\
     & ($CEBB_2$)  & 0.044 & 0.044 & 0.044 & 0.044 & 0.044 & 0.044  &  0.014\\
    \hline
    \multirow{2}{*}{\texttt{cervical-cancer-1}} & ($CEBB_1$)  & 1 & 1 & 1 & 1 & 1  & 1 &  1 \\
    & ($CEBB_2$) & 1 & 1 & 1 & 1 & 1  & 0.999 &  0.999\\
    \hline
    \multirow{2}{*}{\texttt{cervical-cancer-2}} & ($CEBB_1$) & 1 &  0.964 & 0.963 & 0.963 & 0.963  & 0.963  & 0.963 \\
    & ($CEBB_2$) & 1 & 0.765 & 0.598 & 0.539 & 0.486  & 0.494 &  0.493\\
    \hline
    %\texttt{cervical-cancer-3} & 0.058 & 0.058 \\
    %\texttt{cervical-cancer-4} & 0.07 & 0.075\\
    \multirow{2}{*}{\texttt{churn}} & ($CEBB_1$)  & 0.8 & 0.8 & 0.8 & 0.8  &  0.8  & 0.8 &  0.8 \\
    & ($CEBB_2$) & 0.8 & 0.437 & 0.416 & 0.271 & 0.228  & 0.175 & 0.149\\
    \hline
    %\texttt{diabetes} & 0.182 & 0.196  \\
    \multirow{2}{*}{\texttt{german}} & ($CEBB_1$)  & 0.510 & 0.258 & 0.255 & 0.255 & 0.255  & 0.255 &  0.255 \\
     & ($CEBB_2$) & 0.510 & 0.427 & 0.192 & 0.175 & 0.157  & 0.155 &  0.134\\
    \hline
    \multirow{2}{*}{\texttt{heart}} & ($CEBB_1$)  & 0.141 &  0.141 & 0.141 & 0.141 & 0.141 & 0.141  &   0.071 \\
     & ($CEBB_2$) & 0.141 & 0.141 & 0.141 & 0.141 & 0.141 & 0.141  &  0.028 \\
    \hline
    \multirow{2}{*}{\texttt{housing}} & ($CEBB_1$) & 0.114 & 0.114 & 0.114 & 0.114 & 0.114  & 0.057 &  0.057 \\
    & ($CEBB_2$) & 0.114 & 0.114 & 0.114 & 0.114 & 0.114 & 0.114 &  0.035\\
    \hline
    \multirow{2}{*}{\texttt{leukemia}} & ($CEBB_1$)  & 0 & 0 & 0 & 0 & 0  & 0 &  0 \\
    & ($CEBB_2$) & 0 & 0 & 0 & 0 & 0 & 0 &  0\\
    \hline
     \multirow{2}{*}{\texttt{productivity}} & ($CEBB_1$) & 0.354 & 0.354 & 0.352 & 0.180 & 0.180  & 0.180 & 0.180 \\
     & ($CEBB_2$) & 0.354 & 0.354 & 0.156 & 0.137 & 0.123  & 0.103 &  0.081\\
    \hline
     \multirow{2}{*}{\texttt{SRBCT}} & ($CEBB_1$) & 0.063 & 0.063 & 0.063 & 0.063 & 0.063  & 0.063 &  0.063 \\
     & ($CEBB_2$) & 0.063 & 0.063 & 0.063 & 0.063 & 0.063  & 0 & 0\\
    \hline
     \multirow{2}{*}{\texttt{wisconsin}} & ($CEBB_1$) & 0.036 & 0.036 & 0.036 & 0.036  & 0.036 & 0.036 &  0.018 \\
     & ($CEBB_2$) & 0.036 & 0.036 & 0.036 & 0.036 & 0.036  & 0.036 &  0.003\\
    \hline
  \hline
\end{tabular}
}
\caption{Out-of-sample MSE for the positive class probability predictions of each dataset.}\label{fig:CSv11prueba}
\end{table}
\begin{figure}[h!]
\includegraphics[width=13cm]{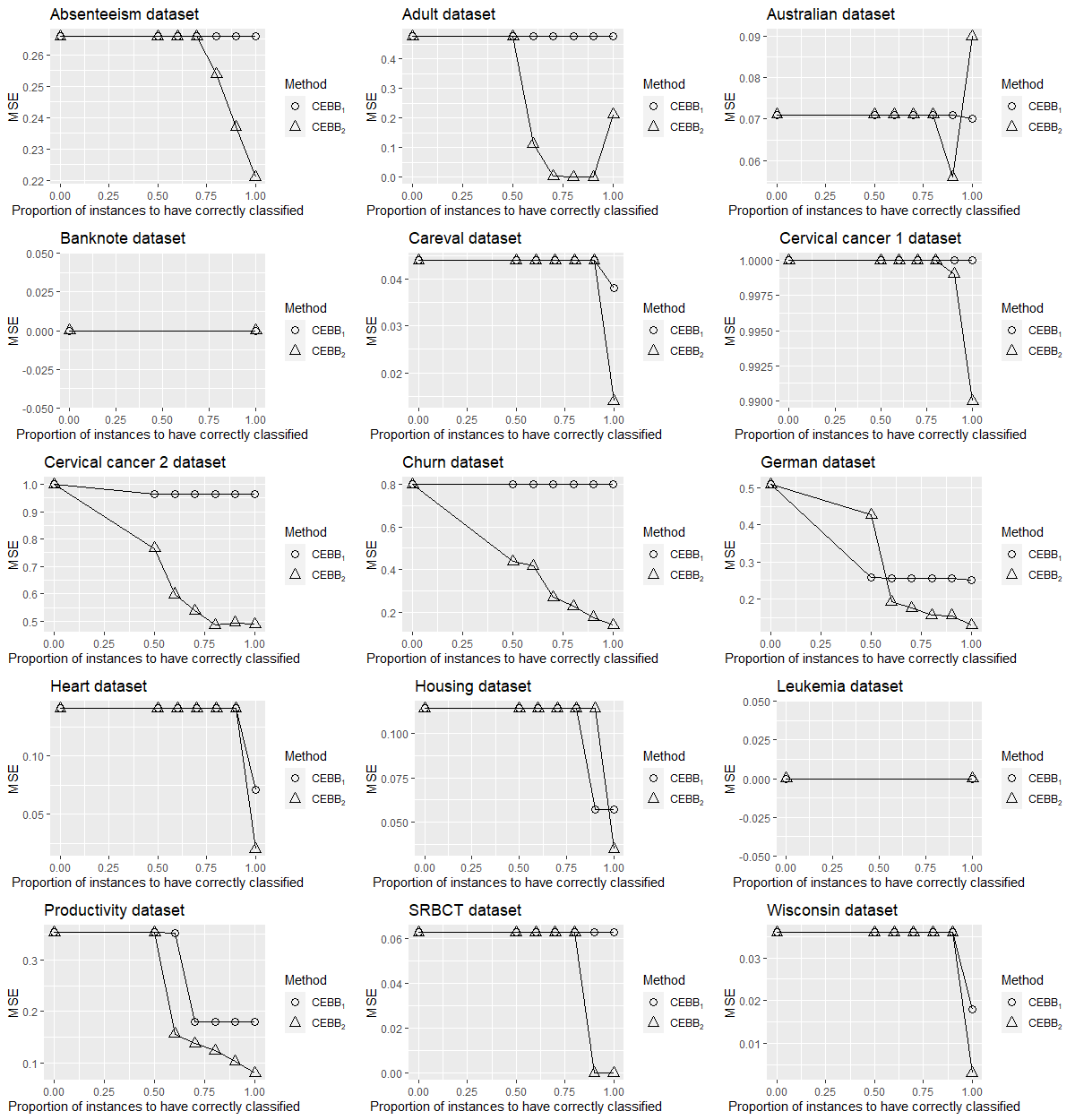}
\caption{Out-of-sample MSE for the positive class probability predictions of each dataset.}\label{fig:controlledPOS1y2}
\end{figure}
However, Table~\ref{fig:CSv11prueba2}{, which contains the MSE for the negative instances when the positive class is controlled,} presents{, in general,} just the opposite pattern.
\begin{table}[h!]
\Rotatebox{0}{
\centering
\begin{tabular}{|cc|ccccccc|}
  \hline
  \multicolumn{2}{|c|}{\text{Dataset (Method) $\backslash$ TPR imposed}} & 0 &  0.5 & 0.6 & 0.7 & 0.8  & 0.9 & 1\\
  \hline
        %\texttt{adult} & 0.163 & 0.158 \\
    \multirow{2}{*}{\texttt{absentism}} & ($CEBB_1$) & 0.088 &  0.088 & 0.088 & 0.088 & 0.088  & 0.044 & 0.044\\
    & ($CEBB_2$) & 0.088 & 0.088 & 0.088 & 0.088 & 0.043  & 0.045 &  0.051\\
    \hline
    \multirow{2}{*}{\texttt{adult}} & ($CEBB_1$) & 0.044 & 0.044 & 0.022 & 0.022 &  0.022  & 0.022 &  0.022 \\
    & ($CEBB_2$) & 0.044 & 0.044 & 0.202 & 0.825 & 0.969  & 0.976 &  0.095\\
    \hline
    \multirow{2}{*}{\texttt{australian}} & ($CEBB_1$) & 0.271 & 0.271 & 0.271 & 0.271 & 0.271  & 0.135 &  0.135 \\
    & ($CEBB_2$) & 0.271 & 0.271 & 0.271 & 0.271 & 0.271  & 0.304 &  0.171 \\
    \hline
    \multirow{2}{*}{\texttt{banknote}} & ($CEBB_1$)  & 0.019 & 0.019 & 0.019 & 0.019 & 0.019  & 0.019 & 0.019 \\
    & ($CEBB_2$) & 0.019 & 0.019 & 0.019 & 0.019 & 0.019  & 0.019 &  0.019\\
    \hline
    %\texttt{cancer-colon} & 0.167 & 0.177 \\
    \multirow{2}{*}{\texttt{careval}} & ($CEBB_1$)   & 0.061 & 0.061 & 0.061 & 0.061 & 0.061 & 0.061 & 0.030 \\
     & ($CEBB_2$) & 0.061 & 0.061 & 0.061 & 0.061 & 0.061 & 0.061  &  0.119\\
    \hline
    \multirow{2}{*}{\texttt{cervical-cancer-1}} & ($CEBB_1$)  & 0.001 & 0.001 & 0.001 & 0.001 & 0.001  &  0.001 &  0.001 \\
    & ($CEBB_2$) & 0.001 & 0.024 & 0.024 & 0.074 & 0.074 & 0.113 & 0.113 \\
    \hline
    \multirow{2}{*}{\texttt{cervical-cancer-2}} & ($CEBB_1$) & 0.008 & 0.004 & 0.004 & 0.004 & 0.004  & 0.004 &  0.004\\
    & ($CEBB_2$) & 0.008 & 0.065 & 0.044 & 0.073 & 0.107  & 0.126 &  0.171 \\
    \hline
    %\texttt{cervical-cancer-3} & 0.058 & 0.058 \\
    %\texttt{cervical-cancer-4} & 0.07 & 0.075\\
    \multirow{2}{*}{\texttt{churn}} & ($CEBB_1$) & 0.008 & 0.008 & 0.004 & 0.004 & 0.004  & 0.004 & 0.004 \\
    & ($CEBB_2$) & 0.008 & 0.049 & 0.060 & 0.076 & 0.044  & 0.065 &  0.097\\
    \hline
    %\texttt{diabetes} & 0.182 & 0.196  \\
    \multirow{2}{*}{\texttt{german}} & ($CEBB_1$)  & 0.051 & 0.051 & 0.051 & 0.051 & 0.051  & 0.051 &  0.051  \\
     & ($CEBB_2$) & 0.051 & 0.093 & 0.125 & 0.145 & 0.207  & 0.252 &  0.421\\
    \hline
    \multirow{2}{*}{\texttt{heart}} & ($CEBB_1$) & 0.207 & 0.207 & 0.207 & 0.207 & 0.207  & 0.207 & 0.071\\
     & ($CEBB_2$) & 0.207 & 0.207 & 0.207 & 0.207 & 0.207 & 0.207 &  0.44\\
    \hline
    \multirow{2}{*}{\texttt{housing}} & ($CEBB_1$)  & 0.046 & 0.046 & 0.046 & 0.046 & 0.046  & 0.046 &  0.046 \\
    & ($CEBB_2$) & 0.046 & 0.046 & 0.046 & 0.046 & 0.046  & 0.046 & 0.296\\
    \hline
    \multirow{2}{*}{\texttt{leukemia}} & ($CEBB_1$)  & 0 & 0 & 0 & 0 & 0 & 0  &  0 \\
    & ($CEBB_2$) & 0 & 0 & 0 & 0 & 0 &  0 & 0\\
    \hline
     \multirow{2}{*}{\texttt{productivity}} & ($CEBB_1$) &  0.119 & 0.119 & 0.136 &  0.140 & 0.140  & 0.140 &  0.140 \\
     & ($CEBB_2$) & 0.119 & 0.119 & 0.142 & 0.159 & 0.201  & 0.234 &  0.294\\
    \hline
     \multirow{2}{*}{\texttt{SRBCT}} & ($CEBB_1$) & 0 & 0 & 0 & 0 & 0  & 0 & 0 \\
     & ($CEBB_2$) & 0 & 0 & 0 & 0 & 0  &  0.137 &  0.090\\
    \hline
     \multirow{2}{*}{\texttt{wisconsin}} & ($CEBB_1$) & 0.006 & 0.006 & 0.006 & 0.006 & 0.006  & 0.006 &  0.006 \\
     & ($CEBB_2$) & 0.006 & 0.006 & 0.006 & 0.006 & 0.006  & 0.006 &  0.017\\
    \hline
\end{tabular}
}
\caption{Out-of-sample MSE for the negative class probability predictions of each dataset.}\label{fig:CSv11prueba2}
\end{table}
{This can be better visualized from Figures~\ref{fig:controlledPOS1y2} (which represents a plot taking the values from Table~\ref{fig:CSv11prueba}) and ~\ref{fig:controledPOSNEG1y2} (that is used as a visualization of the data in Table~\ref{fig:CSv11prueba2}).}
\begin{figure}[h!]
\includegraphics[width=13cm]{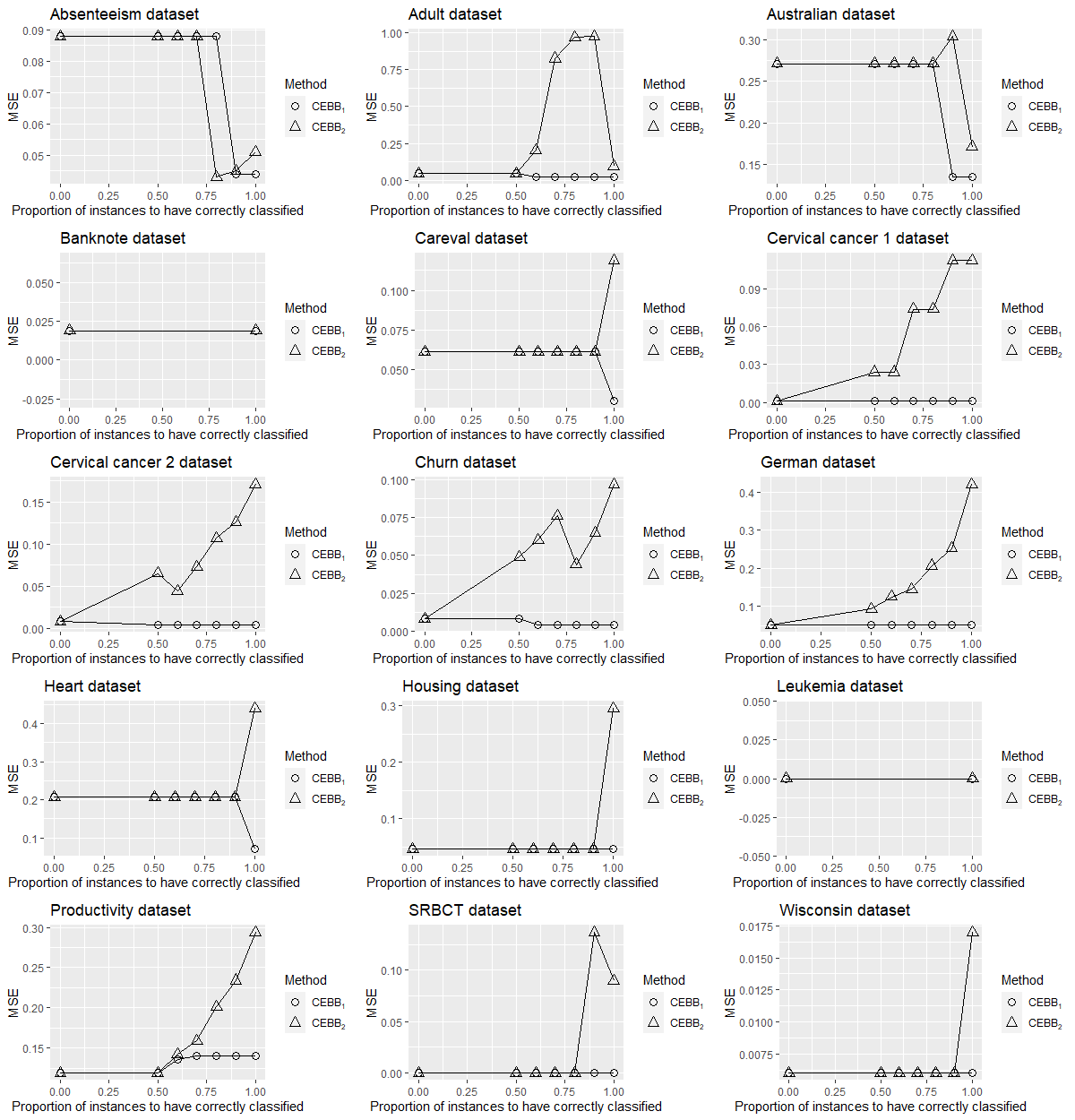}
\caption{Out-of-sample MSE for the negative class probability predictions of each dataset.}\label{fig:controledPOSNEG1y2}
\end{figure}
\textcolor{black}{Here again, some datasets result in almost null MSEs (\texttt{cervical-cancer}, \texttt{leukemia}, \texttt{SRBCT} and \texttt{wisconsin})}. {In view of the tables, and more visually through their respective plots, we can see how, although both techniques are able to control the class of interest, it is $CEBB_2$ that achieves the most significant results in improving the classification of the desired class. As a counterpart to this, the other class (the negative one) suffers further deterioration. Although this is a generalised behaviour, we see that there are some datasets that have some anomalous performance, such as in the case of ``absenteeism'', where the MSEs for both classes are reduced when we try to control the positive class.  We also find the same anomalous behaviour for the datasets ``careval'' and ``heart'', for which we reduce the MSE with both $CEBB_1$ and $CEBB_2$ (more with $CEBB_2$ than with $CEBB_1$), but whose negative class behaves differently with $CEBB_1$ than with $CEBB_2$. In particular, we notice that while the results improve under $CEBB_2$, at the same time the classification of negative instances is also improved to some extent. Going back to the top six most unbalanced datasets, we can see that for ``cervical-cancer-1'' (the most unbalanced dataset of all) we cannot make much improvement in the positive class, although we slightly improve the results under $CEBB_2$. However, for that ``cervical-cancer-1'' dataset we already did better than the other alternatives from the beginning. For the other unbalanced datasets we can see how we are increasingly lowering the MSE obtained for the positive class, especially under the $CEBB_2$ method, which as commented in Section~\ref{subsec:CostSensit} seem to better controlling the class of interest. However, as usual, this improvement in the class of interest is associated with a deterioration in the classification of the other class.}

An important remark to be made concerning the performance of \textit{CEBB$_1$} and \textit{CEBB$_2$} is as follows. The first one seems to be able to improve the sensitivity without damaging too much the specificity or even improving it at the same time, while the second method damages in a more significant way the specificity, but at the same time it leads to better sensitivity values.

%\begin{figure}[h]
%\centering
%\includegraphics[width=10cm]{Slide1.pdf}
%\includegraphics[width=10cm]{Slide3.pdf}
%\includegraphics[width=10cm]{Sliden.pdf}
%    \caption{Original score threshold}
%    \label{materialflowChart}
%\end{figure}

%\begin{figure}[h]
%\centering
%\includegraphics[width=10cm]{Slide12.pdf}\\
%\includegraphics[width=10cm]{Slide32.pdf}
%\includegraphics[width=10cm]{Sliden2.pdf}
 %   \caption{Moved score threshold to improve probabilities in the positive class}
 %   \label{materialflowChart2}
%\end{figure}

\section{Conclusions and further research}\label{sec:Conc}

\textcolor{black}{In this paper we have proposed a procedure to obtain probabilistic outputs for the Support Vector Machines, through point estimates}. Contrary to existing proposals, we present a method that is distribution-free and cost-sensitive. Also, it makes use of not only a single classifier but a weighted average of the scores corresponding to the different classifiers built for the different parameters of the SVM, obtaining more accurate results. \textcolor{black}{The method turns out advantageous for operational business processes as credit scoring or churn prediction, where the class of interest may suffer from imbalancedness.}

Our proposal is compared to some benchmark methodologies. \textcolor{black}{The results show that our approach is comparable or better than such approaches.} Two cost-sensitive alternatives are proposed here. The first one is based on changing the way the probabilities are estimated and the second one proposes to modify the original classifier by a cost-sensitive version. Results for real datasets have been shown, proving the usefulness of the novel approach.

{In our numerical results, the complete datasets have been used without a previous feature selection process aimed to reduce the size of the data. In this respect, we find interesting to consider two future research lines: (1) to analyze the impact of a prior feature selection strategy for high dimensional data, and (2) to study how to embed a variable selection method in the design of the classifier, for example, along the lines of \cite{Sandra1}. Also,} traditional SVM can be used as a basis for addressing multiclass problems. How to extend properly our approach to such multiclass problems is an interesting research avenue which is now under investigation. %Furthermore, up to our knowledge, no prior work has undertaken posterior class probabilities estimation by confidence intervals. How to use the bootstrap approach, which naturally leads to confidence intervals for the scores values, will be object of future study.

\section*{Acknowledgements}
This research is financed by projects EC H2020 MSCA RISE NeEDS Project (Grant agreement ID: 822214), TED2021-130216A-I00 and FJC2021-047209-I (these two funded by MCIN\slash AEI\slash10.13039\slash501100011033 and by European Union ``NextGenerationEU''\slash PRTR) FQM329, P18-FR-2369 and US-1381178 (Junta de Andaluc\'ia, Andaluc\'ia), PR2019-029 (Universidad de C\'adiz) and PID2019-110886RB-I00 (Ministerio de Ciencia, Innovación y Universidades, Spain). The last {three} are cofunded with EU ERD Funds. The authors are thankful for such support.





\newpage

\bibliographystyle{apalike}

\bibliography{paper-arxiv}

\newpage

\section*{Supplementary Material}

{We aim in this section to illustrate the performance of the novel methods regarding a different metric. We have considered the AUC (area under the curve).}

{Table~\ref{tab:ResultsLinearAUC} shows the analogous results to those of Table~\ref{tab:ResultsLinear}, but showing the AUC instead of the MSE.}

\begin{table}[h!]

\centering \small

\begin{tabular}{llllll}

     \hline
  \textit{Dataset} &  \textit{EBB (\text{Best} $\theta$)} & \textit{\text{EBB}} & \textit{Sollich} & \textit{\text{Platt}} & \textit{\text{Tao et al.}} \\
  & & & & & ($r=0$, $\sqrt{10}$, $\sqrt{20}$, $\sqrt{30}$)\\
  \hline
    \texttt{absenteeism} & 0.865 & 0.858 & 0.858 & \textbf{0.881} & 0.821, 0.821, 0.821, 0.821  \\
    \texttt{adult} & 0.741 & 0.741 & \textbf{0.901} & 0.894 & 0.779, 0.801, 0.832, 0.839\\
    \texttt{australian} & 0.878 & 0.875 & 0.897 & \textbf{0.914} & 0.844, 0.848, 0.872, 0.884\\
    \texttt{banknote} & \textbf{1} & \textbf{1} & 0.999  & \textbf{1} & 0.990, 0.994, 0.856, 0.733\\
    %\texttt{cancer-colon} & 0.177 & 0.238 & \textbf{0.156} &   0.174, 0.174, 0.174, 0.174\\
    \texttt{careval} & 0.975 & 0.979 & 0.992 & 0.990 & 0.952, 0.955, \textbf{0.995}, 0.991 \\
    \texttt{cervical-cancer-1} & 0.541 & 0.571 & 0.662 & \textbf{0.688} & 0.503, 0.503, 0.503, 0.503\\
    \texttt{cervical-cancer-2} & 0.516 & 0.51 & 0.531 & \textbf{0.596} & 0.506, 0.506, 0.506, 0.506 \\
    %\texttt{cervical-cancer-3} & 0.058 &  0.236 & \textbf{0.049} & 0.055, 0.055, 0.055, 0.055\\
    %\texttt{cervical-cancer-4} & 0.075 & 0.175 & \textbf{0.061} & 0.075, 0.075, 0.075, 0.075\\
    \texttt{churn} & 0.592 & 0.592 & 0.908 & 0.885 & 0.688, 0.688, \textbf{0.985}, 0.969 \\
    %\texttt{diabetes} & 0.196 & 0.219 & \textbf{0.159} & 0.224, 0.161, 0.210, 0.228\\
    %\texttt{divorce} & 0.059 & 0.124 & \textbf{0.016} & 0.024, 0.024, 0.024, 0.024\\
    \texttt{german} & 0.75 & 0.76 & 0.777 & \textbf{0.795} & 0.661, 0.674, 0.735, 0.790 \\
    \texttt{heart} & 0.837 &  0.838 & 0.914 & 0.897 & 0.868, 0.895, \textbf{0.964}, 0.951 \\
    \texttt{housing} & 0.958 &  \textbf{0.961} & 0.953 & 0.943 & 0.859, 0.914, 0.897, 0.873 \\
    \texttt{leukemia} & \textbf{1} & \textbf{1} & \textbf{1} & 0.992 & 0.983, 0.983, 0.983, 0.983  \\
    \texttt{productivity} & 0.714 &  0.749 & 0.781 & \textbf{0.789} & 0.713, 0.705, 0.681, 0.666 \\
    \texttt{SRBCT} &  \textbf{1} & \textbf{1} & \textbf{1} & \textbf{1} & \textbf{1}, \textbf{1}, \textbf{1}, \textbf{1}  \\
    \texttt{wisconsin} & \textbf{0.996} & \textbf{0.996} & 0.992 & 0.987 & 0.968, 0.981, 0.982, 0.974\\
    \hline

 \end{tabular}

\caption{Out-of-sample AUC obtained when predicting the posterior class probabilities in a linear SVM.}
\label{tab:ResultsLinearAUC}

\end{table}

{We can see that in this case the results vary slightly from those shown when we considered the MSE as the performance metric. However, although the datasets on which each algorithm perform the best have varied, overall the frequency with which each method performs the best have remained the same (\textit{Platt} and $EBB$) or even have increased (as it is the case of \textit{EBB (Best $\theta$)}, \textit{Tao et al}. and \textit{Sollich}).  \\
Looking in detail Table~\ref{tab:ResultsLinearAUC} for the case of ``churn'', it could be thought that our method $EBB$ does perform poorer than the competitors methods. In order to analyze this phenomenom in detail, consider Figure~\ref{fig:Solchurn} that depicts, for 10 random instances, the estimated probabilities under Sollich's approach (in light gray) versus the observed ones (in black colour), which are always either $0$ or $1$. From the Figure, we can see how the probabilities are not fitted correctly, since they take values around 0.5 and thus far from the real values. This fact affects the computation of the MSE value but not that of the AUC and therefore, for this case the AUC classification rate is good in spite of a poor estimation of the probabilities. On the contrary, as shown by Figure~\ref{fig:EBBchurn} (the analogous to Figure~\ref{fig:Solchurn} but for different instances and under the $EBB$ method), our method performs better for capturing the real probabilities since the estimated values are closer to $0$ and $1$. This results in better MSE values but not necessarily in a good AUC performance since a mistake when estimating any of the probabilities (see, e.g., instances \textcolor{black}{7 and 9}) have a significant impact in the AUC computation. This issue is interesting and shows how, in the context of SVMs, the accuracy for the class probabilities is better measured by the MSE than by the AUC.}

\begin{figure}[h!]
\includegraphics[width = 12cm]{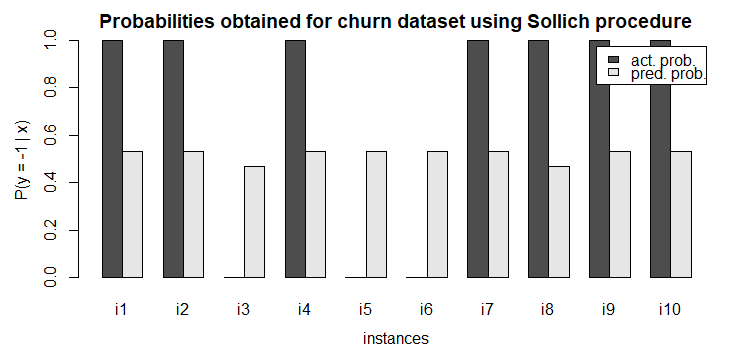}
\caption{Probabilities $P(y=-1 \mid x)$ in 10 random instances (i1 to i10) from ``churn'' dataset. In black, the actual probabilities. In white, the calculated probabilities using \textit{Sollich}.}\label{fig:Solchurn}
\end{figure}

\begin{figure}[h!]
\includegraphics[width = 12cm]{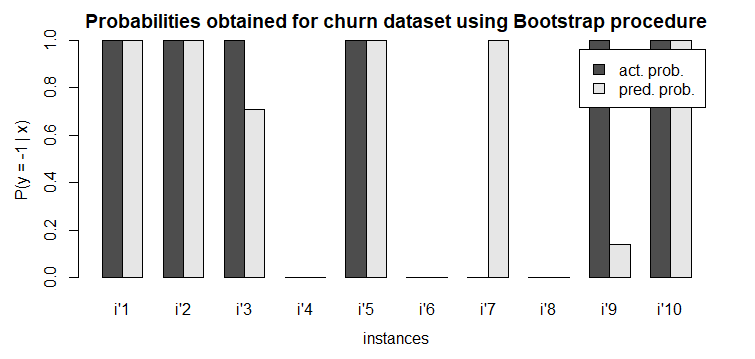}
\caption{Probabilities $P(y=-1 \mid x)$ in 10 random instances (i1 to i10) from ``churn'' dataset. In black, the actual probabilities. In white, the calculated probabilities using \textit{EBB}.}\label{fig:EBBchurn}
\end{figure}

{Finally, we present in Table~\ref{fig:CSv11prueba3AUC} the AUC values obtained for our two cost-sensitive methods ($CEBB_1$ and $CEBB_2$) when we want to control the positive class. As before, the columns indicate the TPR we require to attain.}
\begin{table}[ht]
\Rotatebox{0}{
\centering
\begin{tabular}{|cc|ccccccc|}
  \hline
  \multicolumn{2}{|c|}{\text{Dataset (Method) $\backslash$ TPR imposed}} & 0 &  0.5 & 0.6 & 0.7 & 0.8  & 0.9 & 1\\
  \hline
        %\texttt{adult} & 0.163 & 0.158 \\
    \multirow{2}{*}{\texttt{absentism}} & ($CEBB_1$) & 0.858 & 0.858 & 0.858 &  0.858 & 0.858  & 0.858 & 0.857\\
    & ($CEBB_2$) & 0.858 & 0.858 & 0.858 &  0.858 & 0.857 & 0.866 & 0.873 \\
    \hline
    \multirow{2}{*}{\texttt{adult}} & ($CEBB_1$) & 0.741 & 0.741 & 0.741 & 0.741 & 0.741 & 0.741 & 0.741\\
    & ($CEBB_2$) & 0.741 & 0.741 & 0.847 & 0.803 & 0.767 & 0.52 & 0.802\\
    \hline
    \multirow{2}{*}{\texttt{australian}} & ($CEBB_1$)  &  0.875 & 0.875 & 0.875 & 0.875 & 0.875 & 0.875 & 0.875\\
    & ($CEBB_2$) &   0.875 & 0.875 & 0.875 & 0.875 & 0.875 & 0.857 & 0.829 \\
    \hline
    \multirow{2}{*}{\texttt{banknote}} & ($CEBB_1$)  & 1 & 1 & 1 & 1 & 1 & 1 & 1 \\
    & ($CEBB_2$) & 1 & 1 & 1 & 1 & 1 & 1 & 1 \\
    \hline
    %\texttt{cancer-colon} & 0.167 & 0.177 \\
    \multirow{2}{*}{\texttt{careval}} & ($CEBB_1$)   & 0.979 & 0.979 & 0.979 & 0.979 & 0.979 & 0.979 & 0.979  \\
     & ($CEBB_2$)  & 0.979 & 0.979 & 0.979 & 0.979 & 0.979 & 0.979 & 0.969 \\
    \hline
    \multirow{2}{*}{\texttt{cervical-cancer-1}} & ($CEBB_1$)  & 0.571 & 0.571 & 0.571 & 0.571 & 0.571 & 0.571 & 0.571 \\
    & ($CEBB_2$) & 0.571  & 0.665 & 0.665 & 0.659 & 0.659 & 0.735 & 0.735   \\
    \hline
    \multirow{2}{*}{\texttt{cervical-cancer-2}} & ($CEBB_1$) & 0.51 & 0.51 & 0.506 & 0.516 & 0.51 & 0.51 & 0.51  \\
    & ($CEBB_2$) & 0.51 & 0.53 & 0.534 & 0.645 & 0.622 & 0.6 & 0.607\\
    \hline
    %\texttt{cervical-cancer-3} & 0.058 & 0.058 \\
    %\texttt{cervical-cancer-4} & 0.07 & 0.075\\
    \multirow{2}{*}{\texttt{churn}} & ($CEBB_1$)  &  0.592 &  0.592 &  0.592 &  0.592 &  0.592 &  0.592 &  0.592 \\
    & ($CEBB_2$) & 0.592 & 0.854 &  0.83 & 0.861 & 0.874 & 0.854 & 0.841\\
    \hline
    %\texttt{diabetes} & 0.182 & 0.196  \\
    \multirow{2}{*}{\texttt{german}} & ($CEBB_1$)  & 0.76 & 0.759 & 0.76 & 0.759 & 0.76 & 0.759 & 0.76  \\
     & ($CEBB_2$) & 0.76 & 0.788 & 0.777 & 0.765 & 0.736 &  0.716 & 0.662 \\
    \hline
    \multirow{2}{*}{\texttt{heart}} & ($CEBB_1$)  & 0.838 & 0.838 & 0.838 & 0.838 & 0.838 & 0.837 & 0.837 \\
     & ($CEBB_2$)  & 0.838 & 0.838 & 0.838 & 0.838 & 0.838 & 0.837 & 0.827 \\
    \hline
    \multirow{2}{*}{\texttt{housing}} & ($CEBB_1$) &  0.961 &  0.961 &  0.961 &  0.961 &  0.961 & 0.961 &  0.961  \\
    & ($CEBB_2$)  &  0.961 &  0.961 &  0.961 &  0.961 &  0.961 & 0.924 & 0.932\\
    \hline
    \multirow{2}{*}{\texttt{leukemia}} & ($CEBB_1$)  & 1 & 1 & 1 & 1 & 1 & 1 & 1\\
    & ($CEBB_2$) & 1 & 1 & 1 & 1 & 1 & 1 & 1\\
    \hline
     \multirow{2}{*}{\texttt{productivity}} & ($CEBB_1$) & 0.749 & 0.749 & 0.749 & 0.742 & 0.743 & 0.743 & 0.743\\
     & ($CEBB_2$) & 0.749 & 0.749 & 0.731 & 0.773 & 0.779 & 0.769 & 0.775\\
    \hline
     \multirow{2}{*}{\texttt{SRBCT}} & ($CEBB_1$)  & 1 & 1 & 1 & 1 & 1 & 1 & 1\\
     & ($CEBB_2$) & 1 & 1 & 1 & 1 & 1 & 1 & 1\\
    \hline
     \multirow{2}{*}{\texttt{wisconsin}} & ($CEBB_1$) & 0.996 & 0.996 & 0.996 & 0.996 & 0.996 & 0.996 & 0.996 \\
     & ($CEBB_2$) & 0.996 & 0.996 & 0.996 & 0.996 & 0.996 & 0.996 & 1\\
    \hline
  \hline
\end{tabular}
}
\caption{Out-of-sample AUC for each dataset.}\label{fig:CSv11prueba3AUC}
\end{table}
{If we look at $CEBB_1$, we see how the AUC does not vary practically as we increase the imposed TPR (except for certain errors due to numerical or rounding errors). This is because the TPR here is controlled through a translation, always in parallel, of the separating hyperplane. The AUC is measured in the same way: it is the area under the ROC curve, which is created as a 2D-plot of the ability of a binary classifier system as its discrimination threshold is varied. This variation in the threshold can be seen as the parallel shift of the hyperplane. On the other hand, however, we see a greater variation in the AUC for $CEBB_2$, since here the separating hyperplane we are considering does change completely, as a result of different solutions from different optimization problems (where we change some of the constraints and therefore eliminate feasible solutions that in other cases - for other constraints - could be optimal). Therefore, the differences here are due to a change in the classifier and not to calculation or rounding errors.  Regarding their values, they do not present an homogeneous behavior. On the one hand, we can observe AUCs that go down as we increase the TPR requirements (as in the case of ``german'' dataset). On the other hand, we can see AUCs that go up (as in the case of ``churn'', where we obtained a bad AUC as a base, but then we obtain values comparable to those obtained initially by Sollich, Platt and Tao et al.) when we increase the TPR requisites.}

%%
%% End of file `elsarticle-template-harv.tex'.

\end{document}